\documentclass{article}

\usepackage{multirow} 
\usepackage{graphicx}
\usepackage{amsmath}
\usepackage{algorithm}
\usepackage{booktabs}
\usepackage{subfigure}
\usepackage{wrapfig}
\usepackage{booktabs}

\usepackage{pdfpages}

\usepackage{amsmath,amsfonts,bm}

\usepackage{amssymb,amsthm,mathtools}

\newcommand{\mtd}{\mathrm{mtd}}
\newcommand{\mot}{\mathrm{mot}}
\newcommand{\lin}{\mathrm{lin}}

\newtheorem{assumption}{Assumption}
\newtheorem{lemma}{Lemma}

\newtheorem{remark}{Remark}

\def\eqref#1{equation~\ref{#1}}

\def\1{\bm{1}}

\DeclareMathAlphabet{\mathsfit}{\encodingdefault}{\sfdefault}{m}{sl}
\SetMathAlphabet{\mathsfit}{bold}{\encodingdefault}{\sfdefault}{bx}{n}

 \usepackage[preprint]{neurips_2025}

\usepackage[utf8]{inputenc} 
\usepackage[T1]{fontenc}    
\usepackage{hyperref}       
\usepackage{url}            
\usepackage{booktabs}       
\usepackage{amsfonts}       
\usepackage{nicefrac}       
\usepackage{microtype}      
\usepackage{xcolor}         

\title{Merge-of-Thought Distillation}

\author{
  Zhanming Shen${^{\spadesuit\clubsuit}\footnotemark[1]}$~, 
  Zeyu Qin$^{\clubsuit}$\thanks{$\quad$ Equal Contribution.}~,
  Zenan Huang$^{\clubsuit}$, 
  \textbf{Hao Chen}$^{\spadesuit\clubsuit}$,  \\ 
  \textbf{Jiaqi Hu}$^{\spadesuit\clubsuit}$, 
  \textbf{Yihong Zhuang}$^{\clubsuit}$, 
  \textbf{Guoshan Lu}$^{\clubsuit}$, 
  \textbf{Gang Chen}$^{\spadesuit}$, 
  \textbf{Junbo Zhao}$^{\spadesuit\clubsuit}$\thanks{$\quad$ Corresponding Author.} \\
  $^\spadesuit$Zhejiang University ~$^\clubsuit$Inclusion AI, Ant Group \\ 
  \texttt{\{z.shen, j.zhao\}@zju.edu.cn} \\
}

\begin{document}

\maketitle

\begin{abstract}
Efficient reasoning distillation for long chain-of-thought (CoT) models is increasingly constrained by the assumption of a single oracle teacher, despite the practical availability of multiple candidate teachers and growing CoT corpora. We revisit teacher selection and observe that different students have different “best teachers,” and even for the same student, the best teacher can vary across datasets. Therefore, to unify multiple teachers' reasoning abilities into a student to overcome conflicts among various teachers' supervision, we propose \textbf{Merge-of-Thought Distillation (MoT)}, a lightweight framework that alternates between teacher-specific supervised fine-tuning branches and weight-space merging of the resulting student variants. On competition math benchmarks, using only about 200 CoT samples, applying MoT to a Qwen3-14B student surpasses strong models including Deepseek-R1, Qwen3-32B, and OpenAI-O1, demonstrating substantial gains. Besides, MoT consistently outperforms the best single-teacher distillation, improves general reasoning beyond mathematics while reducing catastrophic forgetting, and shows robustness to distribution-shifted and peer-level teachers. Finally, we have demonstrated MoT possesses consensus CoT by eliminating teacher-specific inductive biases and inter-teacher conflicts while repeatedly reinforcing the learning of consensus reasoning features. These results position MoT as a simple, effective route to efficiently distilling long CoT capabilities from diverse teachers into compact students.
\end{abstract}

\section{Introduction}
As large language models (LLMs) with long chain-of-thought (CoT) capabilities continue to emerge \citep{jaech2024OpenAI,yang2025Qwen3,guo2025deepseek}, reasoning distillation is becoming the key pathway for converting expensive reasoning ability into deployable efficiency. Compared with imitating only final answers, directly supervising the reasoning trajectory enables a smaller student model to learn multi-step solution procedures \citep{luo2025deconstructing,qin2025scaling,guo2025deepseek}. 

Building on these developments, the research focus is shifting from scaling data volume to improving data quality. For example, supervised fine-tuning on only 1{,}000 teacher-distilled samples delivers measurable reasoning gains when paired with test-time compute \citep{muennighoff2025s1}. Likewise, when pretraining already imparts rich mathematical knowledge, a few hundred carefully curated examples can effectively elicit complex reasoning \citep{ye2025limo}. Taken together, these findings indicate that efficiently distilling long CoT trajectories is an effective strategy for training compact models that achieve competitive reasoning accuracy.

\begin{wrapfigure}{r}{0.5\linewidth}
    \centering
    \includegraphics[width=\linewidth]{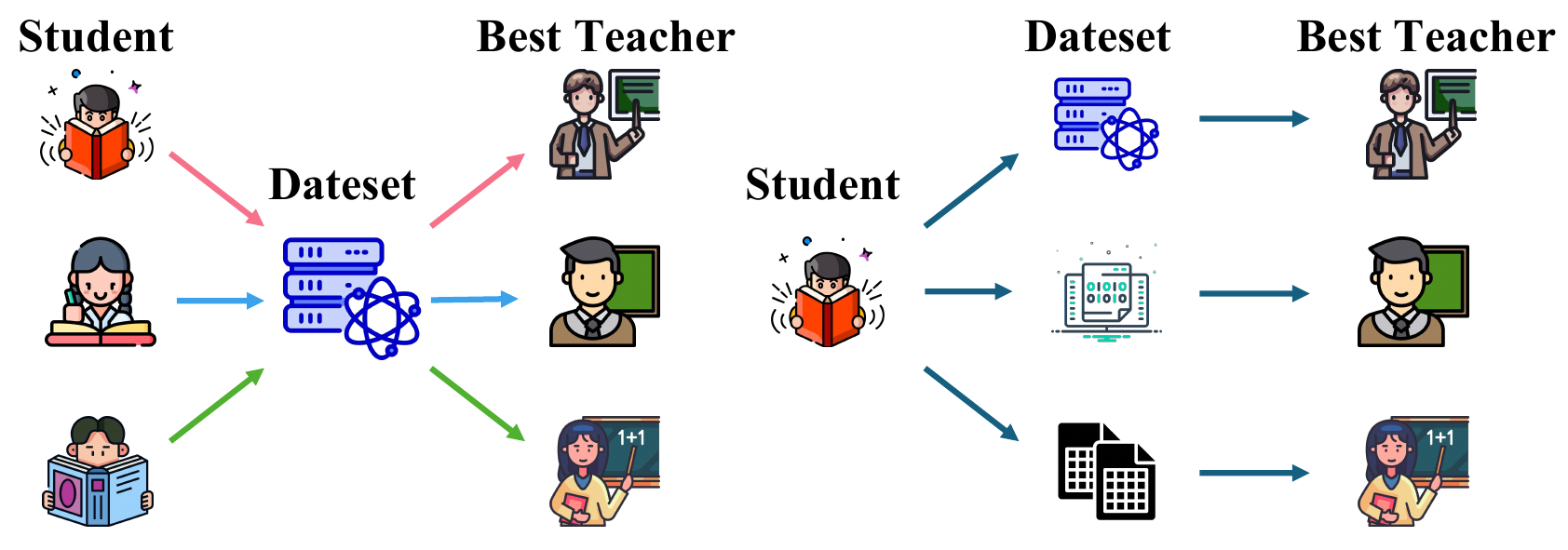}
    \vspace{-5pt}
    \caption{Teacher choice is not universal. Left: different students have different ``best teachers''; right: even for the same student the best teacher can vary across datasets. This observation is empirically confirmed in Table~\ref{tab:best-teacher-by-base}.}
    \vspace{-4pt}
    \label{fig:figure1}
\end{wrapfigure}

However, real-world deployments rarely features a “single oracle teacher.” We often have multiple candidate teacher LLMs and a growing pool of distilled CoT data, giving rise to a basic question: \textbf{\textit{Given a student model, how we pick the most suitable teacher?}} Empirically, teacher choice matters—the teacher can imprint a recognizable “style signature” on the student \citep{chen2025unveiling}; mismatches between teacher and student can weaken the transfer of long CoT skills \citep{wu2025beyond}. As illustrated in Figure~\ref{fig:figure1}, our observations are consistent: different students have different “best teachers,” and even for the same student the best teacher can vary across datasets. Such phenomena challenge the naive assumption that “a bigger/stronger teacher is necessarily better,” prompting us to consider:
\textbf{\textit{Instead of being constrained by a single teacher and the inherent costs of its selection, a more robust and effective paradigm involves aggregating knowledge from multiple teachers.}}


A natural follow-up question is: 
\textit{\textbf{How can we effectively fuse the diverse strengths of multiple teachers?}} The goal is to consolidate their complementary reasoning features into a single student. Long CoTs often accumulate noise and irrelevant content \citep{luo2025through,zhang2025long,li2025feature}. It is unclear whether, in mixed-teacher long-CoT distillation, such noise is amplified through interactions, and how to suppress noise while preserving the consensus features. It suggest that: \textit{\textbf{Diversity of teachers and reasoning paths is an asset—provided we can overcome conflicts among the supervision of various teachers}}.



As an effective technique for overcoming data distribution conflicts, model merging has been widely applied to joint training across diverse domains and tasks \citep{yu2024language,zhou2024metagpt,yadav2024matters}. However, our revisiting analysis also showed that a single Post-hoc merge does not reliably resolve cross-teacher supervision conflicts and unify different teachers' reasoning abilities. These limitations motivate an approach that goes beyond one-shot merging to reconcile heterogeneous teacher signals by \textbf{repeatedly reinforcing the learning of consensus reasoning features}.

%
To this end, we propose \textbf{Merge-of-Thought Distillation (MoT)}: a lightweight framework that alternates between (i) \textbf{teacher-specific branch SFT} and (ii) \textbf{weight-space merging of student variants}. Intuitively, branch SFT internalizes the reasoning style of each teacher into one student; the subsequent parameter-space merge then distills consensus—retaining features reinforced across teachers while suppressing individual accidents and quirks. After multiple iterations, the student progressively condenses into a merged student that reflects multi-teacher consensus reasoning. We found that MoT significantly enhanced reasoning ability of the model and alleviated catastrophic forgetting. In addition, we have experimentally and theoretically demonstrated that \textbf{consensus CoT emerges naturally with MoT}: MoT eliminates teacher-specific inductive biases and inter-teacher conflicts at the token level while repeatedly reinforcing the learning of consensus reasoning features, enabling training in a flatter loss landscape and effective transfer to new student models.



We present, to our knowledge, the \textbf{first systematic study of multi-teacher long CoT co-distillation}:
\begin{enumerate}
    \item We conduct the revisiting analysis of teacher selection under Long CoT distillation setting and find that there is no single best teacher consistently dominant across students or datasets. 
    \item Rather than taking the cost on teacher selection,
    we propose a novel distillation method, \textbf{Merge-of-Thought Distillation (MoT)}, to unify multiple teachers' reasoning abilities into students by overcoming conflicts among the supervision of various teachers. 
    \item Using only about 200 CoT samples, applying MoT to a Qwen3-14B student surpasses strong models including Deepseek-R1, Qwen3-32B, and OpenAI-O1. Besides, MoT consistently outperforms the best single-teacher distillation, improves general reasoning beyond mathematics while reducing catastrophic forgetting, and shows robustness to distribution-shifted and peer-level teachers.
    \item We have demonstrated MoT possesses consensus CoT by eliminating teacher-specific inductive biases and inter-teacher conflicts while repeatedly reinforcing the learning of consensus reasoning features, which enables the model to be trained on a flatter loss landscape and further propagated to new student models.
\end{enumerate}

\section{Related Work}

\paragraph{Long Chain-of-Thought Distillation.}
Research on distilling long chains of thought (CoT) has progressed rapidly \citep{wu2025beyond,guo2025deepseek}. Early work \citep{li2023symbolic} showed that even small models can benefit from teacher CoT prompting and highlighted the importance of varied reasoning chains. 
Subsequent approaches \citep{luo2025deconstructing,feng2024keypoint} further segment and simplify CoTs, employ keypoint weighting, and use progressive distillation to focus on critical tokens. Studies on the key factors of CoT distillation reveal that teacher diversity and rationale granularity often have a greater impact than raw teacher accuracy \citep{chen2025unveiling}. Recent works also show that long‑CoT capability can be bootstrapped with a handful of in‑context examples \citep{pang2025bolt}, distilled as summaries to improve long‑context memory \citep{ma2025recall}, or integrated with vision reasoning using agent‑based approaches \citep{shi2024enhancing}. These findings underscore that long‑CoT distillation not only requires carefully curated examples but also faces challenges such as \textbf{teacher selection}, \textbf{noise amplification} and \textbf{distillation efficiency}. Nevertheless, most existing methods focus on a \textbf{single teacher distillation}; our work instead extends this line of work by fusing multiple teachers’ reasoning abilities into a single student to achieve stronger performance.

\paragraph{Model Merging in LLMs.}
Model merging fuses the parameters of multiple trained models into a single model, which is distinct from output‑level ensembles \citep{yang2024model,tam2024llm}. Empirical studies show that merging tends to balance performance and safety better than mixing data across tasks or languages \citep{yang2025mix,yadav2024matters,yu2024language,jindataless}. More advanced techniques adapt merging to pre‑trained models by disentangling weights into magnitude and direction \citep{yu2024extend}. Other approaches merge checkpoints during pre‑training for faster convergence or use activation importance to retain critical parameters \citep{li2025model,nobari2025activation}.  Model merging has also been applied to combine models with different reasoning strategies and to merge heterogeneous architectures \citep{wu2025unlocking,zhang2024unconstrained}. However, most existing work focuses on merging models specialised for different domains and tasks; by contrast, our approach merges student models distilled by different teachers on the same dataset to \textbf{unify their reasoning abilities} without conflicts among different teachers.

\section{Revisiting Multi-Teacher Long CoT Distillation}
\label{sec:revisiting}

\paragraph{Setup and goals.}
We fine-tune three students from the Qwen3 family (Qwen3-8B / Qwen3-14B / Qwen3-30B-A3B) on two
teacher-distilled math subsets, \textsc{BOBA-200} and \textsc{S1K-200}. We compare three regimes:
(i) \textbf{single-teacher distillation (STD)}, (ii) a \textbf{direct multi-teacher union (MTD)} that mixes all available
teacher-distilled samples, and (iii) a \textbf{one-shot post-hoc weight merge} of students independently
distilled from different teachers. Further dataset/model/training details appear in Sec.~\ref{setup}.
This section has two goals: (1) revisit teacher selection under long CoT distillation and quantify that
the best teacher is student and dataset-dependent; and (2) show that naive MTD or a single
\emph{post-hoc} merge does not reliably resolve cross-teacher supervision conflicts.

\paragraph{Different students have different best teachers.}
Table~\ref{tab:best-teacher-by-base} summarizes, which single-teacher distillation (STD) source achieves the best distillation performance for each base model and dataset. We observe that different students have different best teacher, and \textbf{even for the same student the best teacher can vary across datasets}. This revisiting analysis challenges the naive belief
that a larger/stronger teacher is always better. Details are provided in the Table~\ref{tab:ablation-STD-MTD-MoT}.

\begin{table}[h]
\caption{Best teacher under STD for each base model and dataset.}
\label{tab:best-teacher-by-base}
\begin{center}
\begin{tabular}{lcc}
\toprule
\textbf{Base model} & \textbf{Best teacher on BOBA-200} & \textbf{Best teacher on S1K-200} \\
\midrule
Qwen3-8B        & QWQ          & QWQ \\
Qwen3-14B       & Qwen3-235B   & QWQ \\
Qwen3-30B-A3B   & Qwen3-235B   & Qwen3-235B \\
\bottomrule
\end{tabular}
\end{center}
\end{table}

\paragraph{Simple mixing or one-shot post-hoc merging is insufficient.}
Table~\ref{tab:final-aime-avg} reports final AIME24/25 AVG across two datasets and three student scales. While MTD often improves over the base model, it sometimes lags behind the best per-setting STD especially when the scale of the student model grows. In practice, post-hoc merging behaves similarly to MTD. This means that a straightforward MTD that directly unioning all teachers' distilled samples, and a single post-hoc weight merge of independently distilled students \textbf{do not reliably overcome cross-teacher conflicts or unify heterogeneous reasoning styles, motivating the need for an iterative merge-and-train approach} introduced next in Sec.~\ref{method} to reconcile heterogeneous teacher signals by repeatedly reinforcing the learning of consensus reasoning features.

\begin{table}[h]
\caption{Final AIME24/25 \emph{AVG} under three regimes. MTD denotes naive multi-teacher union.
Best STD denotes best single-teacher for that setting. MTD and Post-hoc weight merge do not reliably overcome cross-teacher conflicts or unify heterogeneous reasoning styles.}
\centering
\setlength{\tabcolsep}{7pt}
\begin{tabular}{llcccc}
\toprule
\textbf{Base Model} & \textbf{Dataset} & \textbf{Baseline} & \textbf{MTD} & \textbf{Best STD} & \textbf{Post-hoc Merge} \\
\midrule
\multirow{2}{*}{Qwen3-8B}
  & BOBA-200 & 71.46 & 72.50 & 71.88        & 73.12 \\
  & S1K-200  & 71.46 & 73.23 & 72.09        & 73.02 \\
\midrule
\multirow{2}{*}{Qwen3-14B}
  & BOBA-200 & 74.59 & 75.94 & 76.98 & 76.98 \\
  & S1K-200  & 74.59 & 76.26 & 76.57        & 76.26 \\
\midrule
\multirow{2}{*}{Qwen3-30B-A3B}
  & BOBA-200 & 75.77 & 76.67 & 78.65 & 78.54 \\
  & S1K-200  & 75.77 & 76.46 & 77.61 & 77.08 \\
\bottomrule
\end{tabular}
\label{tab:final-aime-avg}
\end{table}

\begin{figure}[t]
\centering
\includegraphics[width=\linewidth]{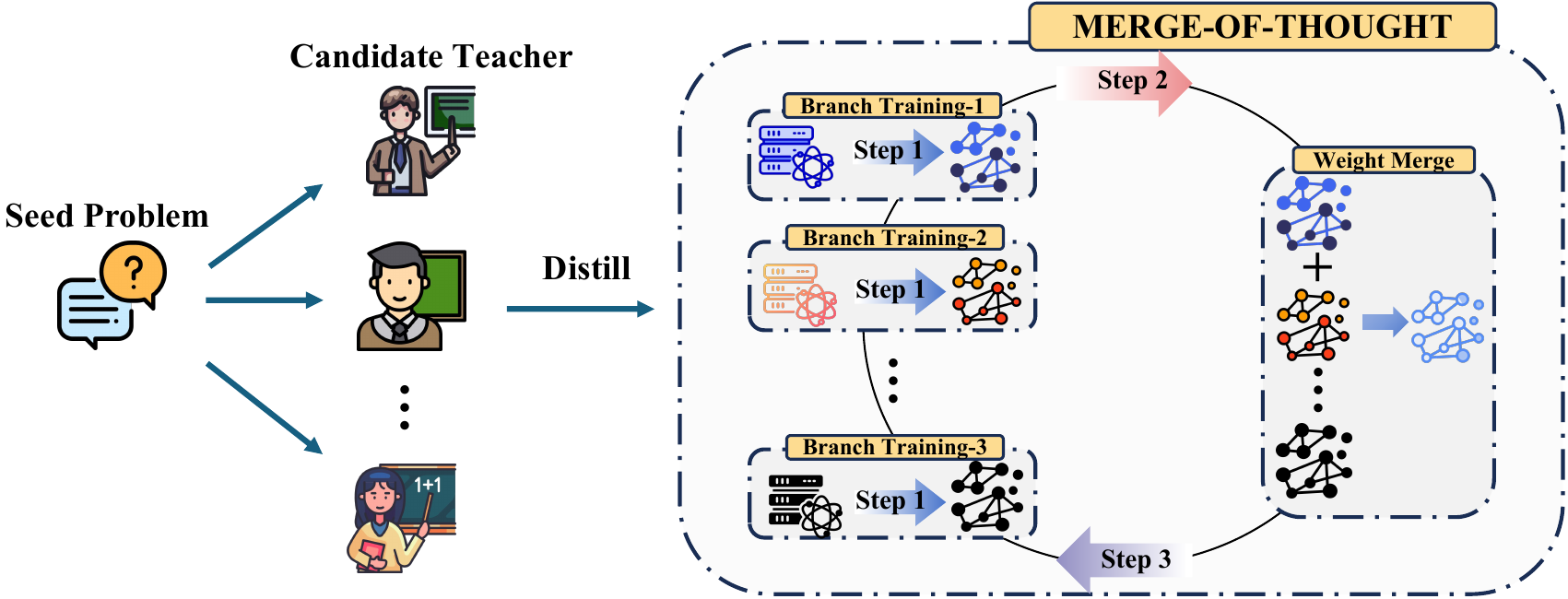} 
\vspace{-15pt}
\caption{Workflow of Merge-of-Thought Distillation (MoT). After the candidate teachers generate the teacher-specific distillation dataset based on the seed problem, the system enters the iterative MoT algorithm process. In each round $t$, we perform three steps:
\textbf{Step~1 (branch training)}: initialize $K$ branches from the current merged student and train each on its teacher-specific distillation dataset $\mathcal{D}^{(k)}$ (Eq.~\ref{eq:sft});
\textbf{Step~2 (weight merge)}: average the branch parameters in weight space to obtain the aggregated model $\theta^{(t)}$ (Eq.~\ref{eq:merge});
\textbf{Step~3 (next-round initialization)}: use $\theta^{(t)}$ as the base initialization for round $t{+}1$.
}
\label{fig:figure2}
\vspace{-8pt}
\end{figure}

\section{Method: Merge-of-Thought Distillation (MoT)}
\label{method}

Our approach assumes access to a base language model, a small set of supervised problems with reference answers, and multiple teacher models. The core idea is to consolidate reasoning signals that are consistent across heterogeneous teacher rationales. MoT alternates between teacher-specific supervised fine-tuning (SFT) branches and weight-space merging, and is performed iteratively. Concretely, MoT consists of two core steps repeated for multiple rounds: \\
1. \textit{Branch training (teacher-specific SFT):} For each teacher, fine-tune a branch of the student on that teacher’s rationales. \\
2. \textit{Weight merge:} Merge branch parameters by averaging to form the next student initialization. 

We detail the setup and these steps below. An overview of the approach is illustrated in Figure~\ref{fig:figure2}.

\subsection{Initialization}

\textbf{Data.} Let $\mathcal{D}=\{(x_i,y_i)\}_{i=1}^N$ be a set of problems $x$ with reference answers $y$. We consider $K$ teacher models. For each input $x$, teacher $\tau_k$ produces a rationale $r^{(k)}$ and a final answer $\hat{y}^{(k)}$. When $y$ is available, we optionally retain only the teacher outputs that match the reference answer, yielding teacher-specific datasets: 
\[
\mathcal{D}^{(k)} \;=\; \{(x_i, r_i^{(k)})\}_{i=1}^{N_k},
\]
which filters out teacher trajectories that do not reach the correct final answer.

\textbf{Model.} Let $m$ denote the student with parameters $\theta$. We initialize from the base model parameters $\theta^{(0)}$ and iterate the MoT procedure for $t=1,\dots,T$ rounds.

\subsection{Teacher-Specific SFT (Branch Training)}

\textbf{Targets.} For each teacher $k$, we train the student to produce the teacher’s rationale:
\[
\text{target}(x; k) \;=\; r^{(k)}.
\]
This choice encourages the student to internalize teacher-specific reasoning patterns, rather than only the short final answer.

\textbf{Objective.} The SFT objective for teacher $k$ is the token-level cross-entropy over the target sequence:
\begin{equation}
\mathcal{L}_{\mathrm{SFT}}^{(k)}(\theta)
\;=\;
\mathbb{E}_{(x, r^{(k)}, y)\sim \mathcal{D}^{(k)}} \;
\sum_{t=1}^{L(x,k)}
-\log p_\theta\bigl(z_t \mid x, z_{<t}\bigr),
\label{eq:sft}
\end{equation}
where $z_{1:L(x,k)}$ tokenizes $\text{target}(x; k)$. In round $t$, we initialize $K$ branches from the current merged model and fine-tune each branch on its teacher’s data:
\[
\theta^{(t,k)} \leftarrow
\operatorname*{arg\,min}_{\theta}\; \mathcal{L}_{\mathrm{SFT}}^{(k)}(\theta)
\quad \text{with init } \theta^{(t-1)}.
\]

\subsection{Weight-Space Merging and Iteration}

After branch training, we merge the $K$ branch parameters by averaging to get the next initialization:
\begin{equation}
\theta^{(t)} \;=\; \frac{1}{K}\sum_{k=1}^K \theta^{(t,k)}.
\label{eq:merge}
\end{equation}
This step consolidates reasoning features that are shared across branches while smoothing out teacher-specific noises. We repeat the two steps—branch training and weight merge—for $T$ rounds, resulting in the final merged model $\theta^{(T)}$. We aim to leverage \emph{model merging} to overcome conflicts among various teachers' supervision and, through continuous merge-and-training iterations, unify different teachers' reasoning abilities and ultimately converge to a consensus reasoning landscape.

\section{Experiments Setup}
\label{setup}

\textbf{Datasets.} We work in a one-question–multiple-answers (1Q–multiA) setting. We use two high-quality open-source mathematical datasets (BOBA \citep{AReaL-boba-Data} and S1K \citep{muennighoff2025s1} as our source datasets. From each source dataset, we sample 200 prompts and denote the resulting subsets as BOBA-200 and S1K-200. For every prompt, we query four teacher models—Qwen3-32B \citep{yang2025Qwen3}, QWQ \citep{team2024qwq}, Deepseek-R1 \citep{guo2025deepseek}, and Qwen3-235B \citep{yang2025Qwen3}. Each teacher generates 16 responses with temperature set to 0.6 and max\_tokens set to 32{,}768. For distillation, we randomly select one correct reasoning path among the 16 as the training label; if none of the 16 responses is correct, we discard that prompt for the corresponding teacher's distillation corpus. We construct two training regimes:  \\
(1) \textbf{Single-Teacher Distillation (STD)}, where we build one distilled corpus per teacher. \\
(2) \textbf{Multi-Teacher Distillation (MTD)}, where we aggregate all available distilled samples from all teachers for each source. \\
The resulting STD and MTD datasets and their sizes are summarized in Table~\ref{tab:distill-all}. Rows with a specific teacher correspond to STD, while rows with “ALL TEACHERS” correspond to MTD.


\textbf{Training Configuration.} We fine-tune Qwen3-8B, Qwen3-14B, and Qwen3-30-A3B \citep{yang2025Qwen3} as base models across all experiments. For MoT, the base model alternates training on each of the four STD corpora for 50 steps and then performs a merge; this constitutes one merge round. We run 5 merge rounds in total and report the best-performing round as the final MoT result; For STD and MTD baselines, to ensure fairness, we train for 250 steps in total and save a checkpoint every 50 steps. We also report the best-performing checkpoint as the final result. More details are provided in the Appendix~\ref{app:ablation1}. We evaluate the capabilities of the model in mathematical reasoning using AIME24 \citep{math-ai2024aime} and AIME25 \citep{aime25}. All AIME scores are 16-run averages.

\begin{table}[h]
\caption{Main results with MoT on BOBA-200 and S1K-200. “/” denotes an item not reported in the corresponding baseline’s source. All AIME scores are 16-run averages.
}
\label{tab:main-MoT-results}
\begin{center}
\scalebox{0.78}{
\begin{tabular}{llccccc}
\toprule[1.2pt]
\multirow{2}{*}{\textbf{Base Model}} & \multirow{2}{*}{\textbf{Configuration}} & \textbf{Annotated} & \multirow{2}{*}{\textbf{AIME24}} & \multirow{2}{*}{\textbf{AIME25}} & \multirow{2}{*}{\textbf{AVG}} & \textbf{AVG} \\
& & \textbf{Examples} & & & & \textbf{Gain} \\
\midrule
\multirow{6}{*}{Qwen3-8B}
  & Base                     & ---  & 75.83 & 67.08 & 71.46 &  - \\
  \cmidrule{2-7}
  & DEER \citep{dai2025s}     & 103K & 76.70 & /     & -     &  - \\
  & S-GRPO \citep{dai2025s}   & 103K & 77.30 & /     & -     &  - \\
  & MathSmith-HC \citep{zhan2025mathsmith}   & 11K  & 76.70 & 70.00 & 73.35 & \footnotesize{\textcolor{teal}{$\uparrow$1.89}} \\
  & \textbf{BOBA-200 + MoT (Ours)}          & 200  & 78.33 & 70.63 & 74.48 & \footnotesize{\textcolor{teal}{$\uparrow$3.02}} \\
  & \textbf{S1K-200 + MoT (Ours)}           & 200  & 77.50 & 71.67 & 74.59 & \footnotesize{\textcolor{teal}{$\uparrow$3.13}} \\
\midrule
\multirow{10}{*}{QWEN2.5-14B}
  & Base                     & ---  & 13.75 & 11.46 & 12.61 &  - \\
  \cmidrule{2-7}
  & GRPO \citep{chen2025spectral}                    & 1K  & 13.33 & 13.13 & 13.23     &  \footnotesize{\textcolor{teal}{$\uparrow$0.62}} \\
  & SPO \citep{chen2025spectral}                     & 1K  & 14.17 & 16.67 & 15.42     &  \footnotesize{\textcolor{teal}{$\uparrow$2.81}} \\
  & RefCritic (SFT) \citep{tang2025refcritic}         & 10K  & 15.20 & 15.00 & 15.10     &  \footnotesize{\textcolor{teal}{$\uparrow$2.49}} \\
  & RefCritic (SFT+RL) \citep{tang2025refcritic}       & 120K & 23.00 & 21.20 & 22.10     &  \footnotesize{\textcolor{teal}{$\uparrow$9.49}} \\
  & Bespoke-Stratos-17k \citep{kou2025data}     & 17K & 20.00 & 13.30 & 16.65     &  \footnotesize{\textcolor{teal}{$\uparrow$4.04}} \\
  & Difficulty-Flipped \citep{kou2025data}        & 17K & 23.00 & 23.30 & 23.15     &  \footnotesize{ \textcolor{teal}{$\uparrow$10.54}} \\
  & Long-CoT \citep{wang2025r1}           & 220K & 30.00 & / & -     &  - \\
  & \textbf{BOBA-200 + MoT (Ours)}           & 200  & 34.17 & 30.00 & 32.09 & \footnotesize{\textcolor{teal}{$\uparrow$19.48}} \\
  & \textbf{S1K-200 + MoT (Ours)}            & 200  & 36.88 & 30.42 & 33.65 & \footnotesize{\textcolor{teal}{$\uparrow$21.04}} \\
\midrule
\multirow{3}{*}{Qwen3-14B}
  & Base                     & ---  & 79.17 & 70.00 & 74.59 &  - \\
  \cmidrule{2-7}
  & \textbf{BOBA-200 + MoT (Ours)}           & 200  & 79.38 & 76.88 & 78.13 & \footnotesize{\textcolor{teal}{$\uparrow$3.54}} \\
  & \textbf{S1K-200 + MoT (Ours)}           & 200  & 81.67 & 75.63 & 78.65 & \footnotesize{\textcolor{teal}{$\uparrow$4.06}} \\
\midrule
\multirow{4}{*}{Qwen3-30B-A3B}
  & Base                     & ---  & 80.63 & 70.90 & 75.77 &  - \\
  \cmidrule{2-7}
  & UloRL-A3B-32k \citep{du2025ulorl}          & /  & / & 73.50 & - & - \\
  & \textbf{S1K-200 + MoT (Ours)}           & 200  & 80.83 & 77.50 & 79.17 & \footnotesize{\textcolor{teal}{$\uparrow$3.40}} \\
  & \textbf{BOBA-200 + MoT (Ours)}          & 200  & 82.92 & 78.33 & 80.63 & \footnotesize{\textcolor{teal}{$\uparrow$4.86}} \\
\midrule
\multirow{1}{*}{Qwen3-32B}
  & Base                     & ---  & 81.46 & 72.08 & 76.77 &  - \\
\midrule
\multirow{1}{*}{Deepseek-R1}
  & Base                     & ---  & 79.80 & 70.00 & 74.90 &  - \\
\midrule
\multirow{1}{*}{OpenAI-O1}
  & Base                     & ---  & 74.30 & 79.20 & 76.75 &  - \\
\midrule
\multirow{1}{*}{OpenAI-O3-MINI}
  & Base                     & ---  & 79.60 & 74.80 & 77.20 &  - \\
\bottomrule[1.2pt]
\end{tabular}
}
\end{center}
\end{table}



\section{Multi-teacher distillation and MoT yield substantial gains}

\subsection{Performance on competition math benchmarks}

\paragraph{Main results.}
To demonstrate the superiority of MoT, we report gains across multiple model scales and compare them against two axes of baselines: (i) larger models like Deepseek-R1, Qwen3-32B and (ii) same-base alternatives trained on methods using substantially larger, differently sourced reasoning datasets. Because the Qwen3 family is very frontier and lacks extensive public baselines, we additionally include results of applying MoT to Qwen2.5-Instruct-14B \citep{team2024qwen2} as a complementary case to test the effectiveness of MoT on 14B scale.

Table~\ref{tab:main-MoT-results} reports the final results of MoT on BOBA-200 and S1K-200. For example, “Qwen3-8B+BOBA-200” denotes Qwen3-8B trained with MoT on BOBA-200 dataset. As shown, \textbf{with only 200 training examples} from either BOBA-200 or S1K-200, MoT lifts Qwen3-8B to match the baseline performance of Qwen3-14B. Moreover, MoT on Qwen3-14B surpasses strong models including Deepseek-R1, Qwen3-32B, and OpenAI-O1, demonstrating substantial gains. In addition, on Qwen2.5-Instruct-14B, MoT's improvements far exceed baselines trained on very large reasoning datasets, reinforcing our claim that \textbf{multi-teacher, consensus-based efficient distillation of long CoT reasoning can yield very substantial performance gains.}

\textbf{Comprehensive Ablations of STD, MTD, and MoT.} To validate the effectiveness of MoT and multi-teacher distillation, we conduct fine-grained ablations: (1) \textbf{STD}: train on each single-teacher distilled dataset (QWQ, Qwen3-32B, Qwen3-235B, Deepseek-R1). (2) \textbf{MTD}: train on the union of all teachers’ distilled samples. (3) \textbf{MoT}: our method that alternates across the four STD corpora with periodic merges. For fairness, all methods save a checkpoint every 50 steps, and we report the best checkpoint; full per-step results are provided in the Appendix~\ref{app:MoT}. 



\begin{table}[h]
\caption{Ablation on STD, MTD, and MoT across settings. AIME scores are 16-run averages.}
\label{tab:ablation-STD-MTD-MoT}
\begin{center}
\scalebox{0.70}{
\begin{tabular}{llccccccccc}
\toprule
 & & \multicolumn{3}{c}{\textbf{Qwen3-8B}} & \multicolumn{3}{c}{\textbf{Qwen3-14B}} & \multicolumn{3}{c}{\textbf{Qwen3-30B-A3B}} \\
\cmidrule(lr){3-5}\cmidrule(lr){6-8}\cmidrule(lr){9-11}
 \textbf{Dataset} & \textbf{Method} & \textbf{AIME24} & \textbf{AIME25} & \textbf{AVG} & \textbf{AIME24} & \textbf{AIME25} & \textbf{AVG} & \textbf{AIME24} & \textbf{AIME25} & \textbf{AVG} \\
\midrule
\multirow{7}{*}{BOBA}
& Baseline                 & 75.83 & 67.08 & 71.46 & 79.17 & 70.00 & 74.59 & 80.63 & 70.90 & 75.77 \\
& MTD (All Teachers)       & 76.04 & 68.96 & 72.50 & 76.46 & 75.42 & 75.94 & 79.38 & 73.96 & 76.67 \\
& STD (QWQ)                & 76.25 & 67.50 & 71.88 & \textbf{79.58} & 73.54 & 76.56 & 79.79 & 75.63 & 77.71 \\
& STD (Qwen3-32B)          & 75.42 & 67.71 & 71.57 & 77.71 & 71.25 & 74.48 & 81.04 & 76.04 & 78.54 \\
& STD (Qwen3-235B)         & 74.58 & 67.92 & 71.25 & 79.17 & 74.79 & 76.98 & 81.88 & 75.42 & 78.65 \\
& STD (Deepseek-R1)        & 67.71 & 60.21 & 63.96 & 74.38 & 67.50 & 70.94 & 78.33 & 68.96 & 73.65 \\
& MoT (ours)               & \textbf{78.33} & \textbf{70.63} & \textbf{74.48} & 79.38 & \textbf{76.88} & \textbf{78.13} & \textbf{82.92} & \textbf{78.33} & \textbf{80.63} \\
\midrule
\multirow{7}{*}{S1K}
& Baseline                 & 75.83 & 67.08 & 71.46 & 79.17 & 70.00 & 74.59 & 80.63 & 70.90 & 75.77 \\
& MTD (All Teachers)       & 75.63 & 70.83 & 73.23 & 79.17 & 73.34 & 76.26 & 78.33 & 74.58 & 76.46 \\
& STD (QWQ)                & 76.04 & 68.13 & 72.09 & 80.21 & 72.92 & 76.57 & \textbf{81.46} & 72.92 & 77.19 \\
& STD (Qwen3-32B)          & \textbf{77.50} & 66.67 & 72.09 & 79.79 & 72.50 & 76.15 & 79.58 & 73.13 & 76.36 \\
& STD (Qwen3-235B)         & 74.38 & 68.54 & 71.46 & 77.08 & 75.41 & 76.25 & 79.17 & 76.04 & 77.61 \\
& STD (Deepseek-R1)        & 70.00 & 61.46 & 65.73 & 73.75 & 62.92 & 68.34 & 78.54 & 70.63 & 74.59 \\
& MoT (ours)               & \textbf{77.50} & \textbf{71.67} & \textbf{74.59} & \textbf{81.67} & \textbf{75.63} & \textbf{78.65} & 80.83 & \textbf{77.50} & \textbf{79.17} \\
\bottomrule
\end{tabular}
}
\end{center}
\end{table}

Results are shown in Table~\ref{tab:ablation-STD-MTD-MoT}. MoT consistently yields the strongest distillation gains in almost all settings, which means that MoT is always superior to the optimal result of the teacher selection method under each setting. This indicates that MoT can sidesteps brittle manual teacher selection by fusing complementary reasoning abilities into a single student.

\begin{figure}[h]
\begin{center}
\newlength{\panelheight}
\setlength{\panelheight}{5cm}
\vspace{-25pt} 
\begin{minipage}[t]{0.24\linewidth}
  \centering
  \begin{minipage}[c][\panelheight][c]{\linewidth}
    \includegraphics[width=\linewidth,height=\panelheight,keepaspectratio]{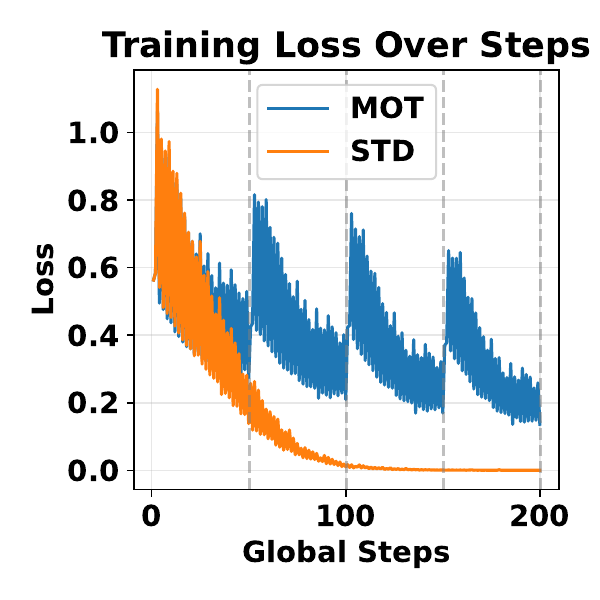} 
  \end{minipage}
  \\[-20pt]
  {\footnotesize (a) S1K: Training loss}
\end{minipage}
\hfill
\begin{minipage}[t]{0.24\linewidth}
  \centering
  \begin{minipage}[c][\panelheight][c]{\linewidth}
    \includegraphics[width=\linewidth,height=\panelheight,keepaspectratio]{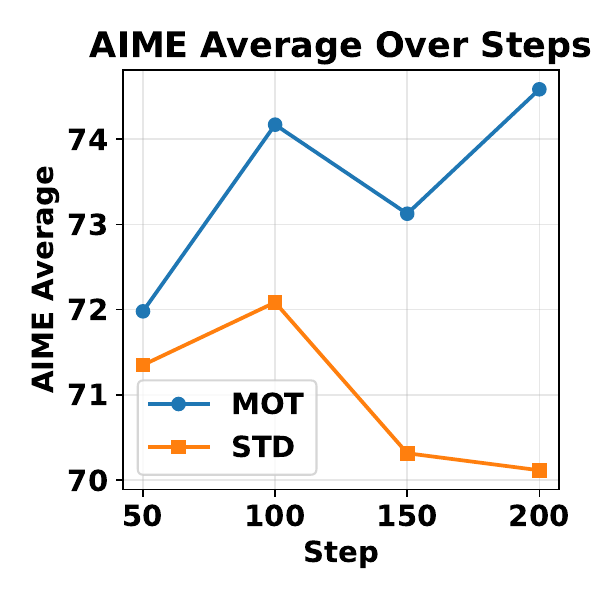} 
  \end{minipage}
  \\[-20pt]
  {\footnotesize (b) S1K: AIME Score}
\end{minipage}
\hfill
\begin{minipage}[t]{0.24\linewidth}
  \centering
  \begin{minipage}[c][\panelheight][c]{\linewidth}
    \includegraphics[width=\linewidth,height=\panelheight,keepaspectratio]{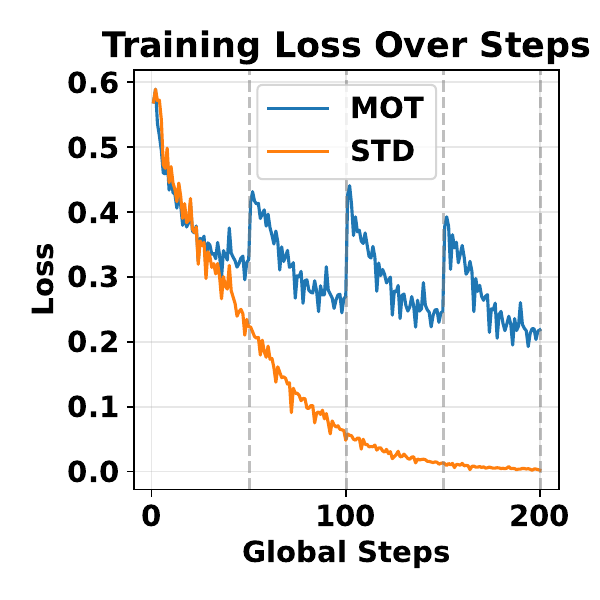}
  \end{minipage}
  \\[-20pt]
  {\footnotesize (c) BOBA: Training loss}
\end{minipage}
\hfill
\begin{minipage}[t]{0.24\linewidth}
  \centering
  \begin{minipage}[c][\panelheight][c]{\linewidth}
    \includegraphics[width=\linewidth,height=\panelheight,keepaspectratio]{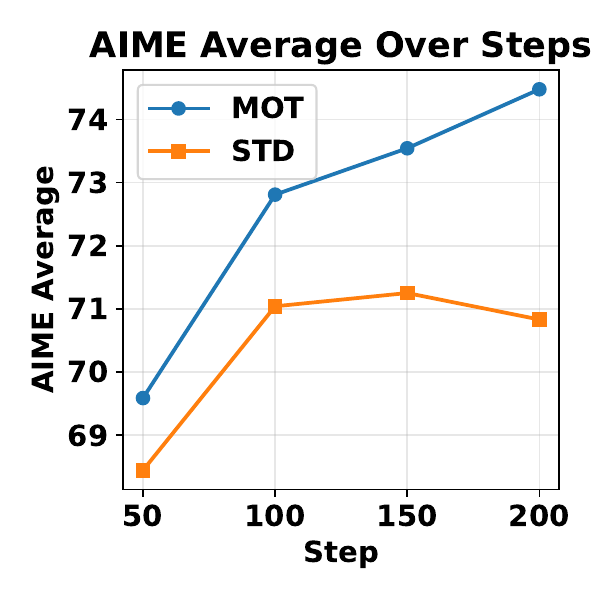}
  \end{minipage}
  \\[-20pt]
  {\footnotesize (d) BOBA: AIME Score}
\end{minipage}

\caption{Qwen3-8B under MoT vs.\ STD (QWQ) on \emph{S1K} and \emph{BOBA}. Panels (a,b): S1K; panels (c,d): BOBA. Left columns show training loss vs.\ steps; right columns show AIME vs.\ steps. All runs log loss at every step on the same QWQ-distilled corpus; AIME is evaluated every 50 steps.}
\vspace{-8pt}
\label{fig:loss-aime-vs-steps-8b}
\end{center}
\end{figure}

\textbf{Training Dynamics: MoT vs.\ Best STD.} We compare Qwen3-8B under MoT and under STD with the best single teacher (QWQ) on both the \emph{S1K} and \emph{BOBA} datasets. We log training loss on the same QWQ-distilled corpora at every step and evaluate AIME score every 50 step. From Figure~\ref{fig:loss-aime-vs-steps-8b}, we observe that MoT achieves substantially higher AIME scores even when its training loss remains much higher than STD’s at the same step. This suggests that in long CoT training, lower loss is not necessarily correlated with stronger reasoning ability. Moreover, MoT \textbf{exhibits a higher performance ceiling and suppresses overfitting}, with STD typically peaking earlier and then degrading while MoT remains stable or continues improving as steps increase.

\subsection{MoT mitigates forgetting and strengthens general reasoning}

To assess whether CoT-style training with MoT affects basic capabilities, we evaluate the final checkpoints trained by MoT and by STD with the per-setting best teacher (Best STD) against the Base models on nine benchmarks: CEVAL (CEV) \citep{seifert2024ceval}, SUPER\_GPQA (SG) \citep{du2025supergpqa}, SIMPLE\_QA (SQ) \citep{wei2024measuring}, IFEVAL (IFE) \citep{zhou2023instruction}, MMLU\_PRO (MP) \citep{wang2024mmlu}, MMLU\_REDUX (MR) \citep{gema2025we}, PhyBench (PB) \citep{meng2024phybench}, LiveCodeBench (LCB) \citep{jain2024livecodebench}, and GPQA-Diamond (GPQA-D) \citep{rein2024gpqa}. We group these benchmarks into three categories: \textbf{catastrophic-forgetting–sensitive tasks}, \textbf{reasoning–knowledge tasks} and \textbf{pure reasoning tasks}. Detailed descriptions of these tasks and MoTivations for using and classifying them for evaluation are provided in the Appendix~\ref{app:benchmarks}.


For each configuration, we report raw scores and summarize the average change versus the Base model within each group: “Avg drop” for catastrophic-forgetting tasks and “Avg gain” for reasoning-knowledge and pure reasoning tasks. We report the results in Table~\ref{tab:catastrophic-forgetting}. Compared with training on the single best teacher, MoT typically yields larger gains on reasoning-knowledge and pure reasoning tasks while incurring smaller declines on catastrophic-forgetting–sensitive tasks. This suggests that MoT not only \textbf{strengthens general reasoning} but also helps \textbf{mitigate catastrophic forgetting}. In Appendix~\ref{app:Task-Type Breakdown}, we provide a more detailed evaluation.

\begin{table}[h]
\caption{Impact of Best STD and MoT on general benchmarks. All scores are 16-run averages.}
\label{tab:catastrophic-forgetting}
\begin{center}
\scalebox{0.63}{
\begin{tabular}{lllcccccccccccc}
\toprule
& & & \multicolumn{4}{c}{\textbf{Catastrophic-forgetting–sensitive tasks}} & \multicolumn{4}{c}{\textbf{Reasoning-knowledge tasks}} & \multicolumn{4}{c}{\textbf{Pure reasoning tasks}} \\
\cmidrule(lr){4-7}\cmidrule(lr){8-11}\cmidrule(lr){12-15}
\textbf{Dataset} & \textbf{Base} & \textbf{Config} & \textbf{CEV} & \textbf{SG} & \textbf{IFE} & \textbf{Avg drop} & \textbf{SQ} & \textbf{MP} & \textbf{MR} & \textbf{Avg gain} & \textbf{PB} & \textbf{LCB} & \textbf{GPQA\text{-}D} & \textbf{Avg gain} \\
\midrule
\multirow{3}{*}{BOBA} & \multirow{3}{*}{8B}
  & Base               & 83.58 & 10.51 & 83.60 & - & 32.31 & 71.42 & 83.21 & - & 20.47 & 55.76 & 57.77 & - \\
& & Best STD              & 83.43 &  9.97 & 81.62 & \footnotesize{\textcolor{purple}{$\downarrow$-0.89}} & 33.88 & 72.00 & 83.68 & \footnotesize{\textcolor{teal}{$\uparrow$0.87}} & 22.85 & 59.88 & 59.85 & \footnotesize{\textcolor{teal}{$\uparrow$2.86}} \\
& & MoT                   & 83.73 & 10.09 & 82.04 & \footnotesize{\textcolor{purple}{$\downarrow$-0.61}} & 34.44 & 73.30 & 84.42 & \footnotesize{\textcolor{teal}{$\uparrow$1.74}} & 24.07 & 58.79 & 60.54 & \footnotesize{\textcolor{teal}{$\uparrow$3.13}} \\
\addlinespace
\multirow{3}{*}{S1K} & \multirow{3}{*}{8B}
  & Base               & 83.58 & 10.51 & 83.60 & - & 32.31 & 71.42 & 83.21 & - & 20.47 & 55.76 & 57.77 & - \\
& & Best STD              & 83.95 & 10.18 & 82.35 & \footnotesize{\textcolor{purple}{$\downarrow$-0.40}} & 32.75 & 72.24 & 85.02 & \footnotesize{\textcolor{teal}{$\uparrow$1.02}} & 22.76 & 59.47 & 56.31 & \footnotesize{\textcolor{teal}{$\uparrow$1.51}} \\
& & MoT                   & 84.32 & 10.15 & 83.51 & \footnotesize{\textcolor{teal}{$\uparrow$0.10}} & 33.56 & 73.01 & 84.95 & \footnotesize{\textcolor{teal}{$\uparrow$1.53}} & 23.37 & 59.58 & 59.53 & \footnotesize{\textcolor{teal}{$\uparrow$2.83}} \\
\midrule
\multirow{3}{*}{BOBA} & \multirow{3}{*}{14B}
  & Base               & 86.78 & 10.76 & 84.69 & - & 32.61 & 75.26 & 85.74 & - & 28.53 & 61.41 & 60.83 & - \\
& & Best STD              & 83.73 & 10.26 & 82.56 & \footnotesize{\textcolor{purple}{$\downarrow$-1.89}} & 32.17 & 74.71 & 86.37 & \footnotesize{\textcolor{purple}{$\downarrow$-0.12}} & 30.61 & 63.21 & 63.79 & \footnotesize{\textcolor{teal}{$\uparrow$2.28}} \\
& & MoT                   & 86.70 & 10.38 & 83.51 & \footnotesize{\textcolor{purple}{$\downarrow$-0.55}} & 32.65 & 75.59 & 86.53 & \footnotesize{\textcolor{teal}{$\uparrow$0.39}} & 30.77 & 63.59 & 64.26 & \footnotesize{\textcolor{teal}{$\uparrow$2.62}} \\
\addlinespace
\multirow{3}{*}{S1K} & \multirow{3}{*}{14B}
  & Base               & 86.78 & 10.76 & 84.69 & - & 32.61 & 75.26 & 85.74 & - & 28.53 & 61.41 & 60.83 & - \\
& & Best STD              & 84.25 & 10.00 & 84.32 & \footnotesize{\textcolor{purple}{$\downarrow$-1.22}} & 32.49 & 76.21 & 86.47 & \footnotesize{\textcolor{teal}{$\uparrow$0.52}} & 30.41 & 63.10 & 63.70 & \footnotesize{\textcolor{teal}{$\uparrow$2.15}} \\
& & MoT                   & 85.66 & 10.45 & 84.42 & \footnotesize{\textcolor{purple}{$\downarrow$-0.57}} & 32.56 & 76.55 & 86.68 & \footnotesize{\textcolor{teal}{$\uparrow$0.73}} & 30.78 & 64.15 & 64.11 & \footnotesize{\textcolor{teal}{$\uparrow$2.76}} \\
\midrule
\multirow{3}{*}{BOBA} & \multirow{3}{*}{30B}
  & Base               & 85.88 & 10.66 & 83.76 & - & 31.68 & 75.26 & 85.81 & - & 28.57 & 61.08 & 59.76 & - \\
& & Best STD              & 84.18 & 10.02 & 80.44 & \footnotesize{\textcolor{purple}{$\downarrow$-1.89}} & 31.52 & 75.96 & 86.04 & \footnotesize{\textcolor{teal}{$\uparrow$0.26}} & 33.31 & 61.34 & 61.81 & \footnotesize{\textcolor{teal}{$\uparrow$2.35}} \\
& & MoT                   & 86.55 & 10.52 & 83.54 & \footnotesize{\textcolor{teal}{$\uparrow$0.10}} & 32.26 & 76.21 & 86.74 & \footnotesize{\textcolor{teal}{$\uparrow$0.82}} & 33.46 & 62.54 & 62.34 & \footnotesize{\textcolor{teal}{$\uparrow$2.98}} \\
\addlinespace
\multirow{3}{*}{S1K} & \multirow{3}{*}{30B}
  & Base               & 85.88 & 10.66 & 83.76 & - & 31.68 & 75.26 & 85.81 & - & 28.57 & 61.08 & 59.76 & - \\
& & Best STD              & 84.62 & 10.04 & 79.74 & \footnotesize{\textcolor{purple}{$\downarrow$-1.97}} & 32.40 & 75.49 & 86.67 & \footnotesize{\textcolor{teal}{$\uparrow$0.60}} & 33.38 & 63.96 & 61.46 & \footnotesize{\textcolor{teal}{$\uparrow$3.13}} \\
& & MoT                   & 86.48 & 10.14 & 82.91 & \footnotesize{\textcolor{purple}{$\downarrow$-0.26}} & 33.19 & 76.49 & 87.28 & \footnotesize{\textcolor{teal}{$\uparrow$1.40}} & 33.40 & 63.92 & 62.53 & \footnotesize{\textcolor{teal}{$\uparrow$3.48}} \\
\bottomrule
\end{tabular}
}
\end{center}
\end{table}

\section{MoT enables Selection-Free CoT Distillation}

\textbf{Ablating a Distribution-Shifted Teacher from MoT: Evidence of Complementarity.} As shown in Table~\ref{tab:ablation-STD-MTD-MoT}, using Deepseek-R1 (R1) as the sole teacher (STD) induces notable performance drops for QWEN bases, indicating a strong distribution shift. To verify that MoT can still leverage useful signals from R1 despite the shift, we ablate R1 from the MoT teacher pool and keep all other settings identical. As shown in Table~\ref{tab:remove-r1}, removing R1 reduces the final MoT performance on BOBA-200 (negative changes), implying that including R1 provides complementary, beneficial supervision that MoT can harness. This proves that MoT can overcome the \textbf{performance degradation} caused by the strong distribution shift teacher and extract \textbf{beneficial common reasoning features} from it. More details are provided in the Appendix~\ref{app:no_r1}.


\begin{wraptable}{r}{0.4\linewidth}
  \vspace{-20pt} 
  \caption{Impact of removing R1 from the MoT teacher pool on BOBA-200.}
  \begin{center}
  \label{tab:remove-r1}
  \begin{tabular}{lc}
    \toprule
    \textbf{Base model}        & \textbf{AVG change} \\
    \midrule
    Qwen3-8B          & -0.62 \\
    Qwen3-14B         & -0.21 \\
    Qwen3-30B-A3B     & -0.42 \\
    \bottomrule
  \end{tabular}
  \end{center}
  \vspace{-\baselineskip} 
\end{wraptable}

\begin{figure}[h]
\begin{center}
\setlength{\panelheight}{5cm} 
\vspace{-31pt} 
\begin{minipage}[t]{0.24\linewidth}
  \centering
  \begin{minipage}[c][\panelheight][c]{\linewidth}
    \includegraphics[width=\linewidth,height=\panelheight,keepaspectratio]{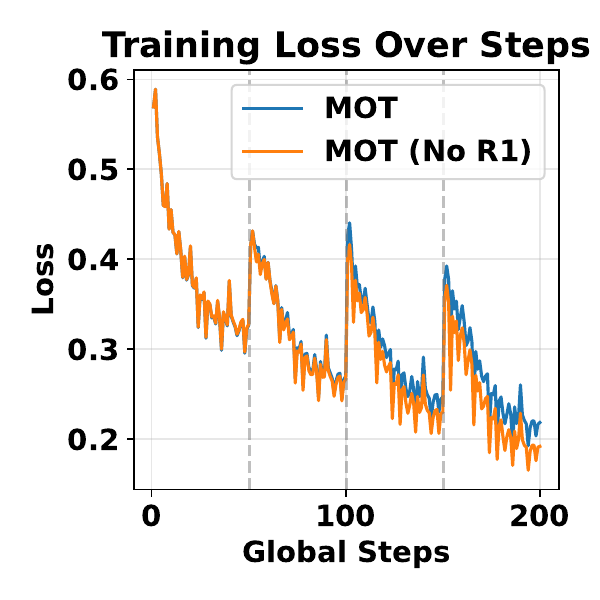}
  \end{minipage}
  \\[-20pt]
  {\footnotesize (a) 8B: Training loss}
\end{minipage}
\hfill
\begin{minipage}[t]{0.24\linewidth}
  \centering
  \begin{minipage}[c][\panelheight][c]{\linewidth}
    \includegraphics[width=\linewidth,height=\panelheight,keepaspectratio]{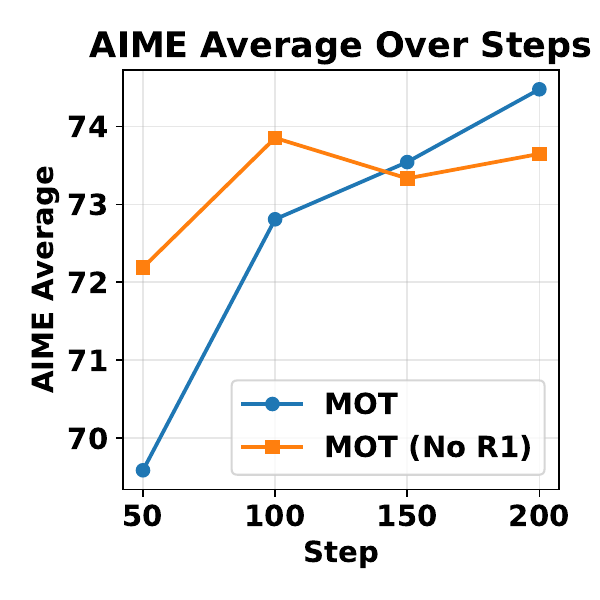}
  \end{minipage}
  \\[-20pt]
  {\footnotesize (b) 8B: AIME Score}
\end{minipage}
\hfill
\begin{minipage}[t]{0.24\linewidth}
  \centering
  \begin{minipage}[c][\panelheight][c]{\linewidth}
    \includegraphics[width=\linewidth,height=\panelheight,keepaspectratio]{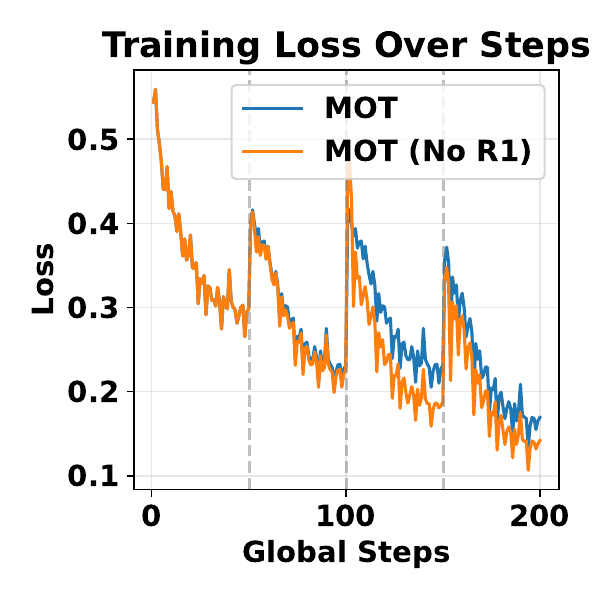} 
  \end{minipage}
  \\[-20pt]
  {\footnotesize (c) 14B: Training loss}
\end{minipage}
\hfill
\begin{minipage}[t]{0.24\linewidth}
  \centering
  \begin{minipage}[c][\panelheight][c]{\linewidth}
    \includegraphics[width=\linewidth,height=\panelheight,keepaspectratio]{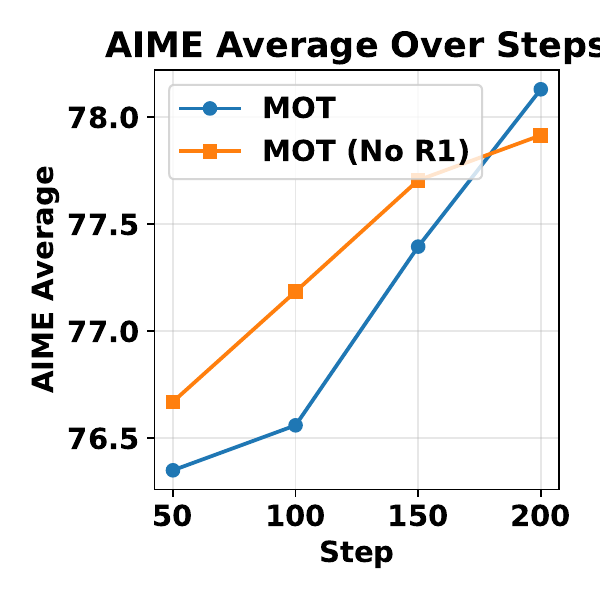} 
  \end{minipage}
  \\[-20pt]
  {\footnotesize (d) 14B: AIME Score}
\end{minipage}

\caption{BOBA dataset: MoT vs.\ MoT without R1 at two scales. Panels (a,b): 8B; panels (c,d): 14B. Left columns show training loss vs.\ steps; right columns show AIME vs.\ steps. All runs log loss at every step on the same QWQ-distilled corpus; AIME is evaluated every 50 steps.}
\label{fig:MoT-vs-MoT-no-r1-8b-14b}
\end{center}
\vspace{-10pt}
\end{figure}

\textbf{Optimization Dynamics with Distribution-Shifted Teacher.} We visualize optimization dynamics on \emph{BOBA} for both \textbf{8B} and \textbf{14B} scales under standard MoT and MoT without R1 (removing the R1 teacher). We log training loss at every step on the same QWQ-distilled corpus and evaluate AIME score every 50 steps (as in our ablation protocol). From Figure~\ref{fig:MoT-vs-MoT-no-r1-8b-14b}, we observe that although the performance of the no-R1 variants converges faster, including R1 \textbf{raises the performance ceiling, delays saturation and reduces post-peak degradation}, suggesting better regularization and a higher training upper bound at both scales. This indicates that even with the distribution-shifted teacher, MoT extracts beneficial common reasoning signals while mitigating teacher-specific noise.

\textbf{Can peer-level models act as teachers?} We find that teacher usefulness extends beyond strictly stronger models: distilling Qwen3-30B-A3B from peer-level QWQ or Qwen3-32B improves performance. Combining peer-level trajectories with MoT boosts results further (Appendix~\ref{app:all_30}).

\section{Consensus CoT emerges naturally with MoT}

\paragraph{Better student is a better teacher.}
To verify that MoT learns higher-quality and more generalizable chains-of-thought (CoT), we conduct a student-as-teacher experiment. Specifically, we take models trained on BOBA-200 under three regimes (Base, Best STD and MoT) and use each \textbf{as a teacher} to re-distill on BOBA-200 for a new student model. As shown in Appendix~\ref{SEC:student-as-teacher}, when the teacher itself is a student trained with MoT, it almost always provides the \textbf{strongest distillation signal}, yielding the best downstream student performance. These results indicate that \textbf{consensus CoT emerges naturally with MoT}: the student learns trajectories that are both stronger and more consistent, and when used as a teacher, this consensus supervision \textbf{transfers} effectively to new students.



\begin{wrapfigure}{r}{0.38\linewidth} 
  \vspace{-12pt} 
  \centering
  \includegraphics[width=\linewidth]{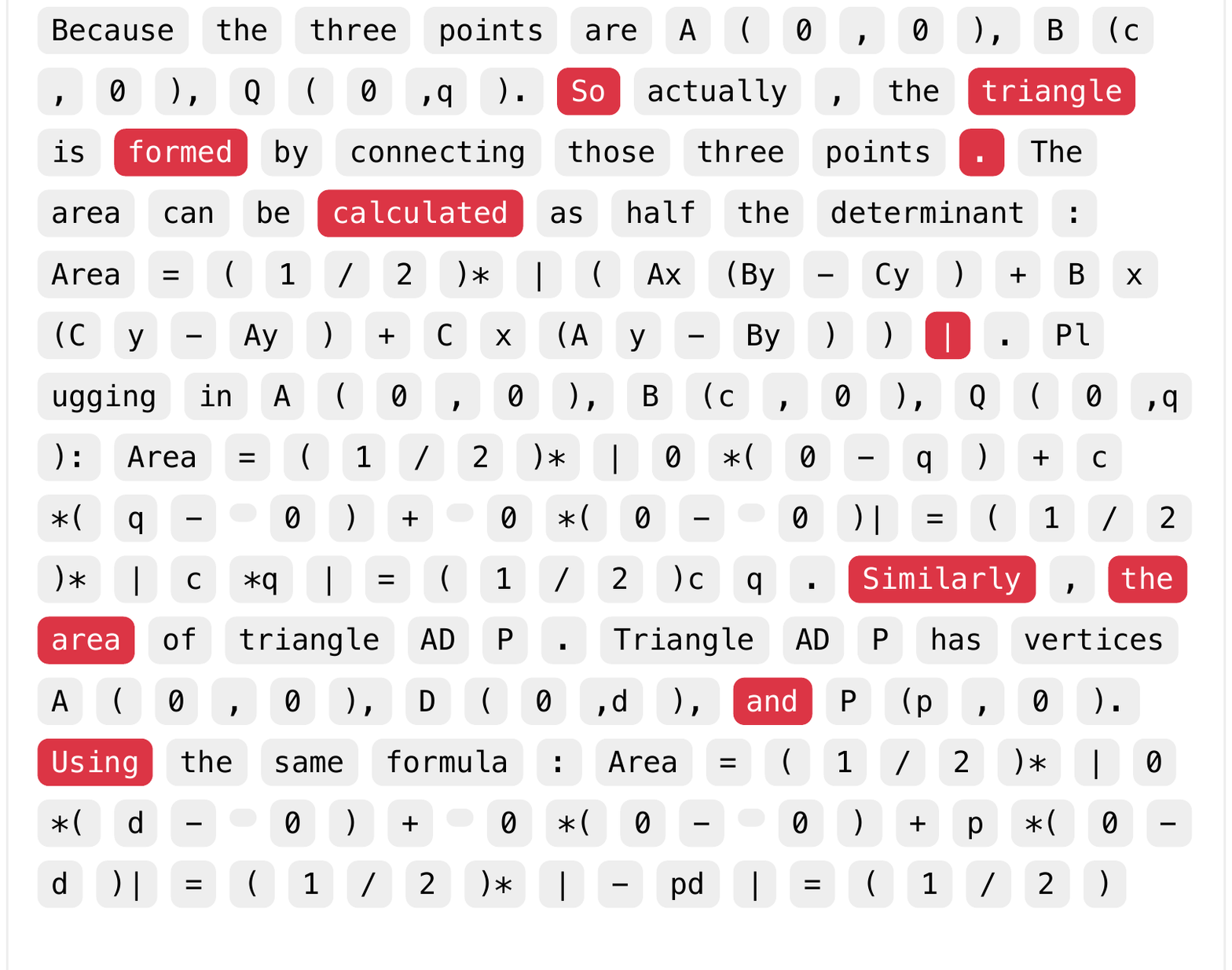}
  \vspace{-20pt}
  \caption{Tokens marked with confidence drops relative to the Base model after MoT.}
  \label{fig:delta_qwq}
  \vspace{-12pt}
\end{wrapfigure}

\paragraph{Token-level evidence for consensus CoT.}
We further probe token-level confidence on QWQ-distilled CoTs. We mark tokens for which the MoT model’s output confidence drops relative to the Base under QWQ teacher’s distilled supervision (Figs.~\ref{fig:delta_qwq}). Strikingly, the marked tokens concentrate on teacher-specific stylistic expressions (driven discourse markers, hedges, and rhetorical flourishes), whereas core derivational tokens (e.g., operators, equations, intermediate results) retain high confidence. This indicates that MoT is essentially \textbf{weakening the learning of inductive bias} of different teachers, while repeatedly \textbf{reinforcing the learning of consensus reasoning ability}. We also detail token-level confidence for MoT and STD(R1) on R1-distilled CoTs in the Appendix Figs.~\ref{fig:tokenconf-r1-three} and Figs.~\ref{fig:tokenconf-r1-std-three}.

\paragraph{MoT mitigates inter-teacher conflicts and trains in a flatter loss landscape.}
We design two complementary evaluations with clear goals: (i) a theoretical “two-bonus” decomposition to test whether expert-wise preconditioning boosts the useful driving term and whether cross-teacher interference is provably reduced—thereby explaining conflict mitigation and flatter updates than MTD; and (ii) a linear mode connectivity probe to check loss landscape flatness of MoT. The “two-bonus” decomposition (Appendix~\ref{sec:analysis-mot-vs-mtd}) shows that expert-wise preconditioning of MoT boosts the driving term while a contractive bound reduces cross-teacher interference, with averaging further shrinking the curvature penalty. Complementarily, a linear mode connectivity probe (Appendix~\ref{subsec:base-to-ckpt-lmc}) indicates that MoT yields markedly smoother loss curves than MTD, indicating flatter regions and reduced sensitivity to teacher noises.

\section{Conclusion}
We presented \textbf{Merge-of-Thought Distillation} (MoT), a lightweight framework that unifies supervision from multiple heterogeneous teachers for long chain-of-thought (CoT) reasoning by alternating teacher-specific SFT with weight-space merging. Revisiting teacher selection shows that different students have different “best teachers,” and even the same student’s best teacher varies across datasets; MoT sidesteps brittle manual selection by fusing complementary reasoning abilities into a single student. With only about \textbf{200} CoT samples, applying MoT to a Qwen3-14B student surpasses Deepseek-R1, Qwen3-32B, and OpenAI-O1. Besides, MoT consistently beats the best single-teacher and naive multi-teacher unions, improves general reasoning while mitigating catastrophic forgetting, and is robust to distribution-shifted and peer-level teachers. Finally, we provide theoretical and empirical evidence that MoT naturally induces a consensus CoT by eliminating teacher-specific inductive biases and inter-teacher conflicts while repeatedly reinforcing the learning of consensus reasoning feature, which enables training in a flatter region of the loss landscape and effective transfer to new student models.




\newpage
\bibliographystyle{plainnat}
\bibliography{ref}

\begin{thebibliography}{50}
\providecommand{\natexlab}[1]{#1}
\providecommand{\url}[1]{\texttt{#1}}
\expandafter\ifx\csname urlstyle\endcsname\relax
  \providecommand{\doi}[1]{doi: #1}\else
  \providecommand{\doi}{doi: \begingroup \urlstyle{rm}\Url}\fi

\bibitem[Chen et~al.(2025{\natexlab{a}})Chen, Li, Li, Chen, and Lin]{chen2025spectral}
Peter Chen, Xiaopeng Li, Ziniu Li, Xi~Chen, and Tianyi Lin.
\newblock Spectral policy optimization: Coloring your incorrect reasoning in grpo.
\newblock \emph{arXiv preprint arXiv:2505.11595}, 2025{\natexlab{a}}.

\bibitem[Chen et~al.(2025{\natexlab{b}})Chen, Sun, Guo, Zhang, Chen, Sun, Su, Pan, Klakow, Li, et~al.]{chen2025unveiling}
Xinghao Chen, Zhijing Sun, Wenjin Guo, Miaoran Zhang, Yanjun Chen, Yirong Sun, Hui Su, Yijie Pan, Dietrich Klakow, Wenjie Li, et~al.
\newblock Unveiling the key factors for distilling chain-of-thought reasoning.
\newblock \emph{arXiv preprint arXiv:2502.18001}, 2025{\natexlab{b}}.

\bibitem[Dai et~al.(2025)Dai, Yang, and Si]{dai2025s}
Muzhi Dai, Chenxu Yang, and Qingyi Si.
\newblock S-grpo: Early exit via reinforcement learning in reasoning models.
\newblock \emph{arXiv preprint arXiv:2505.07686}, 2025.

\bibitem[Du et~al.(2025{\natexlab{a}})Du, Liu, Yang, Chen, and Li]{du2025ulorl}
Dong Du, Shulin Liu, Tao Yang, Shaohua Chen, and Yang Li.
\newblock Ulorl: An ultra-long output reinforcement learning approach for advancing large language models' reasoning abilities.
\newblock \emph{arXiv preprint arXiv:2507.19766}, 2025{\natexlab{a}}.

\bibitem[Du et~al.(2025{\natexlab{b}})Du, Yao, Ma, Wang, Zheng, Zhu, Liu, Liang, Jin, Wei, et~al.]{du2025supergpqa}
Xinrun Du, Yifan Yao, Kaijing Ma, Bingli Wang, Tianyu Zheng, King Zhu, Minghao Liu, Yiming Liang, Xiaolong Jin, Zhenlin Wei, et~al.
\newblock Supergpqa: Scaling llm evaluation across 285 graduate disciplines.
\newblock \emph{arXiv preprint arXiv:2502.14739}, 2025{\natexlab{b}}.

\bibitem[Feng et~al.(2024)Feng, Li, Zhang, Zhou, Yuan, and Wang]{feng2024keypoint}
Kaituo Feng, Changsheng Li, Xiaolu Zhang, Jun Zhou, Ye~Yuan, and Guoren Wang.
\newblock Keypoint-based progressive chain-of-thought distillation for llms.
\newblock In \emph{Proceedings of the 41st International Conference on Machine Learning}, pages 13241--13255, 2024.

\bibitem[Gema et~al.(2025)Gema, Leang, Hong, Devoto, Mancino, Saxena, He, Zhao, Du, Madani, et~al.]{gema2025we}
Aryo~Pradipta Gema, Joshua Ong~Jun Leang, Giwon Hong, Alessio Devoto, Alberto Carlo~Maria Mancino, Rohit Saxena, Xuanli He, Yu~Zhao, Xiaotang Du, Mohammad Reza~Ghasemi Madani, et~al.
\newblock Are we done with mmlu?
\newblock In \emph{Proceedings of the 2025 Conference of the Nations of the Americas Chapter of the Association for Computational Linguistics: Human Language Technologies (Volume 1: Long Papers)}, pages 5069--5096, 2025.

\bibitem[Guo et~al.(2025)Guo, Yang, Zhang, Song, Zhang, Xu, Zhu, Ma, Wang, Bi, et~al.]{guo2025deepseek}
Daya Guo, Dejian Yang, Haowei Zhang, Junxiao Song, Ruoyu Zhang, Runxin Xu, Qihao Zhu, Shirong Ma, Peiyi Wang, Xiao Bi, et~al.
\newblock Deepseek-r1: Incentivizing reasoning capability in llms via reinforcement learning.
\newblock \emph{arXiv preprint arXiv:2501.12948}, 2025.

\bibitem[inclusionAI(2025)]{AReaL-boba-Data}
inclusionAI.
\newblock Areal-boba-data.
\newblock \url{https://huggingface.co/datasets/inclusionAI/AReaL-boba-Data}, 2025.
\newblock URL \url{https://huggingface.co/datasets/inclusionAI/AReaL-boba-Data}.

\bibitem[Jaech et~al.(2024)Jaech, Kalai, Lerer, Richardson, El-Kishky, Low, Helyar, Madry, Beutel, Carney, et~al.]{jaech2024OpenAI}
Aaron Jaech, Adam Kalai, Adam Lerer, Adam Richardson, Ahmed El-Kishky, Aiden Low, Alec Helyar, Aleksander Madry, Alex Beutel, Alex Carney, et~al.
\newblock Openai o1 system card.
\newblock \emph{arXiv preprint arXiv:2412.16720}, 2024.

\bibitem[Jain et~al.(2024)Jain, Han, Gu, Li, Yan, Zhang, Wang, Solar-Lezama, Sen, and Stoica]{jain2024livecodebench}
Naman Jain, King Han, Alex Gu, Wen-Ding Li, Fanjia Yan, Tianjun Zhang, Sida Wang, Armando Solar-Lezama, Koushik Sen, and Ion Stoica.
\newblock Livecodebench: Holistic and contamination free evaluation of large language models for code.
\newblock \emph{arXiv preprint arXiv:2403.07974}, 2024.

\bibitem[Jin et~al.()Jin, Ren, Preotiuc-Pietro, and Cheng]{jindataless}
Xisen Jin, Xiang Ren, Daniel Preotiuc-Pietro, and Pengxiang Cheng.
\newblock Dataless knowledge fusion by merging weights of language models.
\newblock In \emph{The Eleventh International Conference on Learning Representations}.

\bibitem[Kou et~al.(2025)Kou, Tian, Xu, Zeng, and Deng]{kou2025data}
Siqi Kou, Qingyuan Tian, Hanwen Xu, Zihao Zeng, and Zhijie Deng.
\newblock Which data attributes stimulate math and code reasoning? an investigation via influence functions.
\newblock \emph{arXiv preprint arXiv:2505.19949}, 2025.

\bibitem[Li et~al.(2023)Li, Hessel, Yu, Ren, Chang, and Choi]{li2023symbolic}
Liunian~Harold Li, Jack Hessel, Youngjae Yu, Xiang Ren, Kai-Wei Chang, and Yejin Choi.
\newblock Symbolic chain-of-thought distillation: Small models can also" think" step-by-step.
\newblock In \emph{The 61st Annual Meeting Of The Association For Computational Linguistics}, 2023.

\bibitem[Li et~al.(2025{\natexlab{a}})Li, Ma, Yan, Zhang, Liu, Lu, Xu, Chen, Wang, Zhan, et~al.]{li2025model}
Yunshui Li, Yiyuan Ma, Shen Yan, Chaoyi Zhang, Jing Liu, Jianqiao Lu, Ziwen Xu, Mengzhao Chen, Minrui Wang, Shiyi Zhan, et~al.
\newblock Model merging in pre-training of large language models.
\newblock \emph{arXiv preprint arXiv:2505.12082}, 2025{\natexlab{a}}.

\bibitem[Li et~al.(2025{\natexlab{b}})Li, Wang, Yang, Yao, Xiong, and Du]{li2025feature}
Zihao Li, Xu~Wang, Yuzhe Yang, Ziyu Yao, Haoyi Xiong, and Mengnan Du.
\newblock Feature extraction and steering for enhanced chain-of-thought reasoning in language models.
\newblock \emph{arXiv preprint arXiv:2505.15634}, 2025{\natexlab{b}}.

\bibitem[Luo et~al.(2025{\natexlab{a}})Luo, Li, Huang, and Lu]{luo2025through}
Renjie Luo, Jiaxi Li, Chen Huang, and Wei Lu.
\newblock Through the valley: Path to effective long cot training for small language models.
\newblock \emph{arXiv preprint arXiv:2506.07712}, 2025{\natexlab{a}}.

\bibitem[Luo et~al.(2025{\natexlab{b}})Luo, Song, Zhang, Liu, Wang, Chen, Su, and Zheng]{luo2025deconstructing}
Yijia Luo, Yulin Song, Xingyao Zhang, Jiaheng Liu, Weixun Wang, GengRu Chen, Wenbo Su, and Bo~Zheng.
\newblock Deconstructing long chain-of-thought: A structured reasoning optimization framework for long cot distillation.
\newblock \emph{arXiv preprint arXiv:2503.16385}, 2025{\natexlab{b}}.

\bibitem[Ma et~al.(2025)Ma, Fang, Zhang, Zhang, Mi, and Yu]{ma2025recall}
Junyu Ma, Tianqing Fang, Zhisong Zhang, Hongming Zhang, Haitao Mi, and Dong Yu.
\newblock Recall with reasoning: Chain-of-thought distillation for mamba's long-context memory and extrapolation.
\newblock \emph{arXiv preprint arXiv:2505.03320}, 2025.

\bibitem[Math-AI(2024)]{math-ai2024aime}
Math-AI.
\newblock Aime 2024.
\newblock \url{https://huggingface.co/datasets/math-ai/aime24}, 2024.
\newblock URL \url{https://huggingface.co/datasets/math-ai/aime24}.

\bibitem[Math-AI(2025)]{aime25}
Math-AI.
\newblock Aime 2025.
\newblock \url{https://huggingface.co/datasets/math-ai/aime25}, 2025.

\bibitem[Meng et~al.(2024)Meng, Shao, Luo, Wang, Chen, Lu, Yang, Yang, Zhang, Qiao, et~al.]{meng2024phybench}
Fanqing Meng, Wenqi Shao, Lixin Luo, Yahong Wang, Yiran Chen, Quanfeng Lu, Yue Yang, Tianshuo Yang, Kaipeng Zhang, Yu~Qiao, et~al.
\newblock Phybench: A physical commonsense benchmark for evaluating text-to-image models.
\newblock \emph{arXiv preprint arXiv:2406.11802}, 2024.

\bibitem[Muennighoff et~al.(2025)Muennighoff, Yang, Shi, Li, Fei-Fei, Hajishirzi, Zettlemoyer, Liang, Cand{\`e}s, and Hashimoto]{muennighoff2025s1}
Niklas Muennighoff, Zitong Yang, Weijia Shi, Xiang~Lisa Li, Li~Fei-Fei, Hannaneh Hajishirzi, Luke Zettlemoyer, Percy Liang, Emmanuel Cand{\`e}s, and Tatsunori Hashimoto.
\newblock s1: Simple test-time scaling.
\newblock \emph{arXiv preprint arXiv:2501.19393}, 2025.

\bibitem[Nobari et~al.(2025)Nobari, Alimohammadi, ArjomandBigdeli, Srivastava, Ahmed, and Azizan]{nobari2025activation}
Amin~Heyrani Nobari, Kaveh Alimohammadi, Ali ArjomandBigdeli, Akash Srivastava, Faez Ahmed, and Navid Azizan.
\newblock Activation-informed merging of large language models.
\newblock \emph{arXiv preprint arXiv:2502.02421}, 2025.

\bibitem[Pang et~al.(2025)Pang, Dong, Xu, Savarese, Zhou, and Xiong]{pang2025bolt}
Bo~Pang, Hanze Dong, Jiacheng Xu, Silvio Savarese, Yingbo Zhou, and Caiming Xiong.
\newblock Bolt: Bootstrap long chain-of-thought in language models without distillation.
\newblock \emph{arXiv preprint arXiv:2502.03860}, 2025.

\bibitem[Qin et~al.(2025)Qin, Dong, Zhang, Dong, Huang, Yang, Khademi, Zhang, Awadalla, Fung, et~al.]{qin2025scaling}
Zeyu Qin, Qingxiu Dong, Xingxing Zhang, Li~Dong, Xiaolong Huang, Ziyi Yang, Mahmoud Khademi, Dongdong Zhang, Hany~Hassan Awadalla, Yi~R Fung, et~al.
\newblock Scaling laws of synthetic data for language models.
\newblock \emph{arXiv preprint arXiv:2503.19551}, 2025.

\bibitem[Rein et~al.(2024)Rein, Hou, Stickland, Petty, Pang, Dirani, Michael, and Bowman]{rein2024gpqa}
David Rein, Betty~Li Hou, Asa~Cooper Stickland, Jackson Petty, Richard~Yuanzhe Pang, Julien Dirani, Julian Michael, and Samuel~R Bowman.
\newblock Gpqa: A graduate-level google-proof q\&a benchmark.
\newblock In \emph{First Conference on Language Modeling}, 2024.

\bibitem[Seifert et~al.(2024)Seifert, Schl{\"o}tterer, et~al.]{seifert2024ceval}
Christin Seifert, J{\"o}rg Schl{\"o}tterer, et~al.
\newblock Ceval: A benchmark for evaluating counterfactual text generation.
\newblock In \emph{Proceedings of the 17th International Natural Language Generation Conference}, pages 55--69, 2024.

\bibitem[Shi et~al.(2024)Shi, Di, Chen, and Xie]{shi2024enhancing}
Yudi Shi, Shangzhe Di, Qirui Chen, and Weidi Xie.
\newblock Enhancing video-llm reasoning via agent-of-thoughts distillation.
\newblock \emph{arXiv preprint arXiv:2412.01694}, 2024.

\bibitem[Tam et~al.(2024)Tam, Li, Yadav, Gabrielsson, Zhu, Greenewald, Yurochkin, Bansal, Raffel, and Choshen]{tam2024llm}
Derek Tam, Margaret Li, Prateek Yadav, Rickard~Br{\"u}el Gabrielsson, Jiacheng Zhu, Kristjan Greenewald, Mikhail Yurochkin, Mohit Bansal, Colin Raffel, and Leshem Choshen.
\newblock Llm merging: Building llms efficiently through merging.
\newblock In \emph{NeurIPS 2024 Competition Track}, 2024.

\bibitem[Tang et~al.(2025)Tang, Xiang, Yu, Yu, Lin, Lu, Han, Sun, and Lin]{tang2025refcritic}
Qiaoyu Tang, Hao Xiang, Le~Yu, Bowen Yu, Hongyu Lin, Yaojie Lu, Xianpei Han, Le~Sun, and Junyang Lin.
\newblock Refcritic: Training long chain-of-thought critic models with refinement feedback.
\newblock \emph{arXiv preprint arXiv:2507.15024}, 2025.

\bibitem[Team(2024{\natexlab{a}})]{team2024qwen2}
Qwen Team.
\newblock Qwen2. 5: A party of foundation models, 2024{\natexlab{a}}.

\bibitem[Team(2024{\natexlab{b}})]{team2024qwq}
Qwen Team.
\newblock Qwq: Reflect deeply on the boundaries of the unknown.
\newblock \emph{Hugging Face}, 2024{\natexlab{b}}.

\bibitem[Wang et~al.(2025)Wang, Shen, Yao, Huang, Liu, Tan, Huang, Zhang, and Tao]{wang2025r1}
Yibo Wang, Li~Shen, Huanjin Yao, Tiansheng Huang, Rui Liu, Naiqiang Tan, Jiaxing Huang, Kai Zhang, and Dacheng Tao.
\newblock R1-compress: Long chain-of-thought compression via chunk compression and search.
\newblock \emph{arXiv preprint arXiv:2505.16838}, 2025.

\bibitem[Wang et~al.(2024)Wang, Ma, Zhang, Ni, Chandra, Guo, Ren, Arulraj, He, Jiang, et~al.]{wang2024mmlu}
Yubo Wang, Xueguang Ma, Ge~Zhang, Yuansheng Ni, Abhranil Chandra, Shiguang Guo, Weiming Ren, Aaran Arulraj, Xuan He, Ziyan Jiang, et~al.
\newblock Mmlu-pro: A more robust and challenging multi-task language understanding benchmark.
\newblock \emph{Advances in Neural Information Processing Systems}, 37:\penalty0 95266--95290, 2024.

\bibitem[Wei et~al.(2024)Wei, Karina, Chung, Jiao, Papay, Glaese, Schulman, and Fedus]{wei2024measuring}
Jason Wei, Nguyen Karina, Hyung~Won Chung, Yunxin~Joy Jiao, Spencer Papay, Amelia Glaese, John Schulman, and William Fedus.
\newblock Measuring short-form factuality in large language models.
\newblock \emph{arXiv preprint arXiv:2411.04368}, 2024.

\bibitem[Wu et~al.(2025{\natexlab{a}})Wu, Yao, Liu, Liu, Fu, Han, Li, Zhen, Zhong, and Yuan]{wu2025unlocking}
Han Wu, Yuxuan Yao, Shuqi Liu, Zehua Liu, Xiaojin Fu, Xiongwei Han, Xing Li, Hui-Ling Zhen, Tao Zhong, and Mingxuan Yuan.
\newblock Unlocking efficient long-to-short llm reasoning with model merging.
\newblock \emph{arXiv preprint arXiv:2503.20641}, 2025{\natexlab{a}}.

\bibitem[Wu et~al.(2025{\natexlab{b}})Wu, Jiang, Li, Zhai, Liu, Hao, Liu, Yang, Xie, Gu, et~al.]{wu2025beyond}
Xiaojun Wu, Xiaoguang Jiang, Huiyang Li, Jucai Zhai, Dengfeng Liu, Qiaobo Hao, Huang Liu, Zhiguo Yang, Ji~Xie, Ninglun Gu, et~al.
\newblock Beyond scaling law: A data-efficient distillation framework for reasoning.
\newblock \emph{arXiv preprint arXiv:2508.09883}, 2025{\natexlab{b}}.

\bibitem[Yadav et~al.(2024)Yadav, Vu, Lai, Chronopoulou, Faruqui, Bansal, and Munkhdalai]{yadav2024matters}
Prateek Yadav, Tu~Vu, Jonathan Lai, Alexandra Chronopoulou, Manaal Faruqui, Mohit Bansal, and Tsendsuren Munkhdalai.
\newblock What matters for model merging at scale?
\newblock \emph{arXiv preprint arXiv:2410.03617}, 2024.

\bibitem[Yang et~al.(2025{\natexlab{a}})Yang, Li, Yang, Zhang, Hui, Zheng, Yu, Gao, Huang, Lv, et~al.]{yang2025Qwen3}
An~Yang, Anfeng Li, Baosong Yang, Beichen Zhang, Binyuan Hui, Bo~Zheng, Bowen Yu, Chang Gao, Chengen Huang, Chenxu Lv, et~al.
\newblock Qwen3 technical report.
\newblock \emph{arXiv preprint arXiv:2505.09388}, 2025{\natexlab{a}}.

\bibitem[Yang et~al.(2024)Yang, Shen, Guo, Wang, Cao, Zhang, and Tao]{yang2024model}
Enneng Yang, Li~Shen, Guibing Guo, Xingwei Wang, Xiaochun Cao, Jie Zhang, and Dacheng Tao.
\newblock Model merging in llms, mllms, and beyond: Methods, theories, applications and opportunities.
\newblock \emph{arXiv preprint arXiv:2408.07666}, 2024.

\bibitem[Yang et~al.(2025{\natexlab{b}})Yang, Jin, Tang, Shen, Zhu, Chen, Zhao, Wang, Cui, Zhang, et~al.]{yang2025mix}
Jinluan Yang, Dingnan Jin, Anke Tang, Li~Shen, Didi Zhu, Zhengyu Chen, Ziyu Zhao, Daixin Wang, Qing Cui, Zhiqiang Zhang, et~al.
\newblock Mix data or merge models? balancing the helpfulness, honesty, and harmlessness of large language model via model merging.
\newblock \emph{arXiv preprint arXiv:2502.06876}, 2025{\natexlab{b}}.

\bibitem[Ye et~al.(2025)Ye, Huang, Xiao, Chern, Xia, and Liu]{ye2025limo}
Yixin Ye, Zhen Huang, Yang Xiao, Ethan Chern, Shijie Xia, and Pengfei Liu.
\newblock Limo: Less is more for reasoning.
\newblock \emph{arXiv preprint arXiv:2502.03387}, 2025.

\bibitem[Yu et~al.(2024{\natexlab{a}})Yu, Yu, Yu, Huang, and Li]{yu2024extend}
Le~Yu, Bowen Yu, Haiyang Yu, Fei Huang, and Yongbin Li.
\newblock Extend model merging from fine-tuned to pre-trained large language models via weight disentanglement.
\newblock \emph{arXiv preprint arXiv:2408.03092}, 2024{\natexlab{a}}.

\bibitem[Yu et~al.(2024{\natexlab{b}})Yu, Yu, Yu, Huang, and Li]{yu2024language}
Le~Yu, Bowen Yu, Haiyang Yu, Fei Huang, and Yongbin Li.
\newblock Language models are super mario: Absorbing abilities from homologous models as a free lunch.
\newblock In \emph{Forty-first International Conference on Machine Learning}, 2024{\natexlab{b}}.

\bibitem[Zhan et~al.(2025)Zhan, Lai, Lu, Lin, Yang, and Tang]{zhan2025mathsmith}
Shaoxiong Zhan, Yanlin Lai, Ziyu Lu, Dahua Lin, Ziqing Yang, and Fei Tang.
\newblock Mathsmith: Towards extremely hard mathematical reasoning by forging synthetic problems with a reinforced policy.
\newblock \emph{arXiv preprint arXiv:2508.05592}, 2025.

\bibitem[Zhang et~al.(2025)Zhang, Xiao, and Cao]{zhang2025long}
Ruiqi Zhang, Changyi Xiao, and Yixin Cao.
\newblock Long or short cot? investigating instance-level switch of large reasoning models.
\newblock \emph{arXiv preprint arXiv:2506.04182}, 2025.

\bibitem[Zhang et~al.(2024)Zhang, He, Zhang, Fu, Zhou, Sang, Hong, Yang, Wang, Yuan, et~al.]{zhang2024unconstrained}
Yiming Zhang, Baoyi He, Shengyu Zhang, Yuhao Fu, Qi~Zhou, Zhijie Sang, Zijin Hong, Kejing Yang, Wenjun Wang, Jianbo Yuan, et~al.
\newblock Unconstrained model merging for enhanced llm reasoning.
\newblock \emph{arXiv preprint arXiv:2410.13699}, 2024.

\bibitem[Zhou et~al.(2023)Zhou, Lu, Mishra, Brahma, Basu, Luan, Zhou, and Hou]{zhou2023instruction}
Jeffrey Zhou, Tianjian Lu, Swaroop Mishra, Siddhartha Brahma, Sujoy Basu, Yi~Luan, Denny Zhou, and Le~Hou.
\newblock Instruction-following evaluation for large language models.
\newblock \emph{arXiv preprint arXiv:2311.07911}, 2023.

\bibitem[Zhou et~al.(2024)Zhou, Song, Wang, and Chen]{zhou2024metagpt}
Yuyan Zhou, Liang Song, Bingning Wang, and Weipeng Chen.
\newblock Metagpt: Merging large language models using model exclusive task arithmetic.
\newblock \emph{arXiv preprint arXiv:2406.11385}, 2024.

\end{thebibliography}
\appendix
\newpage




\section{Theoretical Analysis}

\label{sec:analysis-mot-vs-mtd}

In this section, we will provide a detailed theoretical analysis to explain the advantages of MoT over MTD in addressing conflicts and mitigating forgetting issues.

Our analysis is based on a comparison of \textbf{the gradient update processes of MoT and MTD}.

\paragraph{Preliminary.}

We approximate the model update for each expert by using second-order Taylor expansion:
\[
\ell_k(\theta)
\approx \ell_k(\theta_{t-1})
+ g_k^\top(\theta-\theta_{t-1})
+ \tfrac12\,(\theta-\theta_{t-1})^\top H_k(\theta-\theta_{t-1}),
\]
where $
g_k = \nabla \ell_k(\theta_{t-1}), \ \ 
H_k = \nabla^2 \ell_k(\theta_{t-1})$, and $\ell_k(\theta)$ is the loss function for expert $k$ evaluated at point $\theta$. We also define the mixture gradient and Hessian as weighted sums of the individual gradients and Hessians:
\[
\bar{g} = \sum_k \alpha_k g_k, \quad \bar{H} = \sum_k \alpha_k H_k,
\]
where $\alpha_k \geq 0$ and $\sum_k \alpha_k = 1$ are the weights assigned to each expert.

Each branch performs $E_k$ steps of gradient descent with a stepsize $\eta$ starting from $\theta_{t-1}$. Based on second-order Taylor expansion, we have $ \theta_{k,E} \;=\; \theta_{t-1} - P_k\, g_k,\ \  P_k \;=\; \eta \sum_{e=0}^{E-1} (I - \eta H_k)^{e} $, where $P_k$ is the ``preconditioner" used in each branch's local optimization process.

We also have the below closed-form solution for preconditioner:
\[
P_k \;=\; s_{E_k}(H_k), \qquad
s_{E}(\lambda) = \frac{1-(1-\eta \lambda)^E}{\lambda}
\;=\; \eta \sum_{e=0}^{E-1} (1-\eta \lambda)^e,
\]
where $s_E(\lambda)$ represents the effective step size along the direction defined by the eigenvalue $\lambda$ of the Hessian matrix $H_k$. 

The expression for $s_E(\lambda)$ can be derived by considering the update rule for gradient descent in the presence of a Hessian, where each step of gradient descent applies a scaling factor depending on the eigenvalue $\lambda$ of the Hessian matrix at each iteration. For large $E$ or small $\eta \lambda$, $s_E(\lambda)$ approximates the inverse of the eigenvalue $\lambda$, leading to more efficient updates along lower-curvature directions.

Hence, the branch displacement for expert $k$ is given by:
\[
\delta_k = -P_k g_k = -s_{E_k}(H_k) g_k,
\]
and the MoT merge, which aggregates the displacements from all experts, is:
\[
\Delta = \sum_k \alpha_k \delta_k = -\sum_k \alpha_k P_k g_k.
\]

For the MTD, which also runs $E$ local steps at the same anchor point, the preconditioner is defined as:
\[
P_{\mtd} = s_E(\bar H),
\]
where $\bar{H}$ is the weighted sum of the Hessians of all experts, and the $E$-step update is:
\[
-\; P_{\mtd} \bar{g} = -s_E(\bar{H}) \bar{g}.
\]
Here, $P_{\mtd}$ is the preconditioner used for the mixture of experts, and $\bar{g}$ is the mixture gradient.

\begin{assumption}[Local quadratic \& stable steps]
\label{asp:quadratic}
Each $\ell_k$ is $C^2$ in a neighborhood $\mathcal N$ of $\theta_{t-1}$.
Let $H_k=\nabla^2\ell_k(\theta_{t-1})$ and $L_{\max}=\max_k \lambda_{\max}(H_k)$.
We choose a stepsize $\eta\in(0,\,2/L_{\max})$ and run $E_k\!\ge\!1$ local steps
whose iterates remain in $\mathcal N$.
\end{assumption}

\paragraph{Two bonuses on the linear part.}
The one-round improvement under the quadratic surrogate
$F_Q(\delta)=\bar g^\top\delta+\tfrac12\delta^\top \bar H\,\delta$ splits into a
\emph{linear} ``driving'' term and a \emph{quadratic} penalty.
For the linear term we have the following variance-type decompositions:
\begin{align}
\label{eq:mot-lin}
\underbrace{\Big\|\sum_k \alpha_k P_k g_k\Big\|^2}_{\text{MoT linear}}
&= \underbrace{\sum_k \alpha_k\,\langle g_k,P_k g_k\rangle}_{{\text{expert-wise preconditioning}}}
\;-\;
\underbrace{\tfrac12\sum_{i,j}\alpha_i\alpha_j\|P_i g_i-P_j g_j\|^2}_{{I_{\mot}\ \ge0}}\!,\\
\label{eq:mtd-lin}
\underbrace{\|\bar g\|_{P_{\mtd}}^2}_{\text{mtd linear}}
&= \underbrace{\sum_k \alpha_k\,\langle g_k,P_{\mtd} g_k\rangle}_{\text{single preconditioner}}
\;-\;
\underbrace{\sum_k \alpha_k\,\|g_k-\bar g\|_{P_{\mtd}}^2}_{{I_{\mtd}(P_{\mtd})\ \ge0}}\!,
\end{align}
where $\|x\|_{M}^2=x^\top M x$.
Subtracting \eqref{eq:mtd-lin} from \eqref{eq:mot-lin} yields the \emph{two-bonus}
difference
\begin{equation}
\label{eq:two-bonus}
\begin{aligned}
\Delta_{\lin}
&\coloneqq
\Big\|\sum_{k} \alpha_k P_k g_k\Big\|^{2}
-\|\bar g\|_{P_{\mtd}}^{2} \\
&=
\underbrace{\sum_{k}\alpha_k\langle g_k,(P_k-P_{\mtd})g_k\rangle}_{\textcolor{blue}{\text{(A') preconditioning gain}}}
\;+\;
\underbrace{I_{\mtd}(P_{\mtd})-I_{\mot}}_{\textcolor{blue}{\text{(B') interference mitigation}}}.
\end{aligned}
\end{equation}

\paragraph{When is (A') $\ge 0$?} 
\begin{lemma}[Monotonicity of $s_E$]
For any fixed $E\ge1$ and $\eta>0$,
$s_E(\lambda)=\eta\sum_{e=0}^{E-1}(1-\eta\lambda)^e$ is strictly decreasing in $\lambda$
on $(0,2/\eta)$.
\end{lemma}

If $H_k$ and $\bar H$ are (approximately) simultaneously diagonalizable, then
$D_k\!\coloneqq\!\langle g_k,(P_k\!-\!P_{\mtd})g_k\rangle
=\|g_k\|^2\sum_r w_{k,r}\big(s_E(\lambda_{k,r})-s_E(\bar\lambda_r)\big)$,
with weights $w_{k,r}=\frac{(q_r^\top g_k)^2}{\|g_k\|^2}$.
Hence $D_k\ge0$ whenever most weight lies on directions where
$\lambda_{k,r}\le \bar\lambda_r$.
Aggregating with $\alpha_k$ gives (A')$\ge0$.


\paragraph{When is (B') $\ge 0$?  A contractive bound on interference.}
Let $\mathcal S=\mathrm{span}\{g_i-g_j\}_{i,j}$ be the disagreement subspace.

\textbf{Assumption (direction-wise contraction on $\mathcal S$).}
There exists $\rho\in(0,1]$ such that on $\mathcal S$ one of the following equivalent
conditions holds:
\begin{enumerate}
  \item[\emph{(Coord.)}] $H_k$ and $\bar H$ are (approximately) simultaneously diagonalizable
  on $\mathcal S$ with eigenbasis $\{q_r\}$; let $p_{k,r}=s_E(\lambda_{k,r})$ and
  $p_{\mtd,r}=s_E(\bar\lambda_r)$. For all $r$ with $q_r\in\mathcal S$,
  \[
  \max_i\, p_{i,r} \;\le\; \rho\, p_{\mtd,r}.
  \]
  \item[\emph{(Basis-free)}] For all $v\in\mathcal S$ and all $k$,
  \[
  \|P_k v\|^2 \;\le\; \rho^2\, \|v\|_{P_{\mtd}}^{2}\qquad
  (\text{i.e., } v^\top P_k^\top P_k v \le \rho^2\, v^\top P_{\mtd} v).
  \]
\end{enumerate}
The above is natural on high-curvature/disagreement directions because
$s_E(\lambda)$ is decreasing in $\lambda$: along directions where at least one expert
has directional curvature no smaller than the mixture (a common empirical pattern),
its preconditioning coefficient is smaller, yielding stronger contraction.

\medskip
Under this assumption we have
\begin{equation}
\label{eq:interf-bound}
I_{\mot}
=\tfrac12\sum_{i,j}\alpha_i\alpha_j\|P_i g_i-P_j g_j\|^2
\ \le\ \rho^2\,\tfrac12\sum_{i,j}\alpha_i\alpha_j\|g_i-g_j\|_{P_{\mtd}}^2
\ =\ \rho^2\,I_{\mtd}(P_{\mtd}).
\end{equation}
Hence $(\mathrm{B}')=I_{\mtd}(P_{\mtd})-I_{\mot}\ \ge\ (1-\rho^2)\,I_{\mtd}(P_{\mtd})\ge 0$.

\paragraph{Implicit shrinkage from averaging enters the quadratic penalty.}
With $\Delta=-\sum_k \alpha_k P_k g_k$, the quadratic penalties satisfy
\begin{align}
R_{\mot}&=\tfrac12\,\Delta^\top \bar H\,\Delta\ \le\
\tfrac12\,\lambda_{\max}(\bar H)\Big(\sum_k \alpha_k\|P_k g_k\|^2\ \underbrace{-\ I_{\mot}}_{\text{shrinkage from averaging}}\Big),\\
R_{\mtd}&=\tfrac12\,\eta^2\,\bar g^\top \bar H\,\bar g\
\le\ \tfrac12\,\lambda_{\max}(\bar H)\Big(\sum_k \alpha_k\|P_{\mtd} g_k\|^2\ -\ I_{\mtd}(P_{\mtd})\Big).
\end{align}
Note the \emph{minus} interference terms, showing that averaging contracts the update
norm and directly reduces the curvature penalty.

\paragraph{Net one-round advantage.}
Combining \eqref{eq:two-bonus}--\eqref{eq:interf-bound} and the penalty bounds yields
\[
\underbrace{\Delta_{\mot}-\Delta_{\mtd}}_{\text{MoT minus MTD}}
\ \gtrsim\ 
\underbrace{\sum_k \alpha_k\langle g_k,(P_k-P_{\mtd})g_k\rangle}_{\text{(A')}}
\;+\;
\underbrace{(1-\rho^2)\,I_{\mtd}(P_{\mtd})}_{\text{(B')}}
\;-\;
\tfrac12\,\lambda_{\max}(\bar H)\cdot \Big[\ \cdots\ \Big],
\]
where $[\cdots]$ gathers the (usually small in the stable regime) difference of squared
update norms.
Thus, under gradient/curvature heterogeneity and stable steps, MoT enjoys a larger
linear driving term (A') and smaller interference (B'), while averaging further cuts
the quadratic penalty.

\paragraph{Special case $E=1$ (for reference).}
Then $P_k=P_{\mtd}=\eta I$, and \eqref{eq:two-bonus} reduces to the familiar
two-term decomposition
\[
\underbrace{\sum_k \alpha_k f_k \|g_k\|^2 - \eta\|\bar g\|^2}_{f_k=\eta}
= \underbrace{0}_{\text{(A)}}\;+\;
\underbrace{\eta\sum_k \alpha_k \|g_k-\bar g\|^2}_{\text{(B) variance bonus}}.
\]

\begin{remark}[Implicit proximal effect (Mitigating Forgetting)]
The matrix series identity $P_k=\eta\sum_{e=0}^{E_k-1}(I-\eta H_k)^e$ shows a
direction-dependent shrink toward the anchor; in each eigendirection $\lambda$ the
effective step is $s_E(\lambda)$, \textbf{larger for low curvature and smaller for high
curvature}, explaining MoT's stability without explicit proximal terms.
\end{remark}

\begin{remark}[Unified Improvements (Mitigating Conflicts)]
A positive value for {\color{blue} both bonus terms} indicates that MoT reduces gradient interference and produces a larger effective update, thereby improving optimization progress. 
\end{remark}

\section{Limitations}
(1) We currently merge branches via simple uniform parameter averaging; future work will explore alternative merge strategies.

(2) Beyond AIME24/25, there is a lack of sufficiently challenging math benchmarks, which limits evaluation depth on high-difficulty mathematical reasoning.

(3) Baseline results in the main results are taken from the original papers/reports because many baselines do not release code/models or disclose key training details like data curation or key hyperparameters. Consequently, they were not re-evaluated under a unified, consistent evaluation configuration, which may affect strict comparability.

\section{Better student is a better teacher}
\label{SEC:student-as-teacher}


\begin{table}[H]
\vspace{-6pt}
\centering
\caption{Student-as-teacher distillation on BOBA-200. Teachers are base model or student models obtained with Best-STD/MoT. We report raw scores on reasoning benchmarks mentioned earlier.}

\label{tab:student-as-teacher}
\begin{center}
\scalebox{0.68}{
\begin{tabular}{lllcccccc}
\toprule
\textbf{Teacher model} & \textbf{Student model} & \textbf{Teacher Config} & \textbf{AIME24} & \textbf{AIME25} & \textbf{PhyBench} & \textbf{LiveCodeBench} & \textbf{GPQA-Diamond} & \textbf{AVG} \\
\midrule
\multirow{3}{*}{Qwen3-14B} & \multirow{3}{*}{Qwen3-8B}
  & Base       & 74.17 & 67.08 & 23.06 & \textbf{58.98} & 57.80 & 56.22 \\
& & Best STD            & 75.21 & 64.17 & 23.74 & 56.74 & 58.33 & 55.64 \\
& & MoT                   & \textbf{75.63} & \textbf{68.96} & \textbf{24.28} & 58.83 & \textbf{59.22} & \textbf{57.38} \\
\addlinespace
\multirow{3}{*}{Qwen3-30B-A3B} & \multirow{3}{*}{Qwen3-14B}
  & Base (Vanilla)        & 79.17 & 68.96 & 28.31 & 61.41 & 61.65 & 59.90 \\
& & Best STD             & 77.08 & \textbf{71.88} & 29.40 & \textbf{63.36} & 61.87 & 60.72 \\
& & MoT                   & \textbf{80.00} & 71.67 & \textbf{29.63} & 62.99 & \textbf{62.69} & \textbf{61.40} \\
\bottomrule
\end{tabular}
}
\end{center}
\end{table}

\section{Probing loss-landscape flatness via base-to-checkpoint interpolation: MoT vs.\ MTD}
\label{subsec:base-to-ckpt-lmc}

\textbf{Setup and purpose.}
To assess how stably a trained model sits in parameter space, we probe \textbf{loss-landscape flatness}
via linear mode connectivity (LMC) between the \emph{base} model and the final trained checkpoint
(from either MTD or our MoT). For $\lambda\in[0,1]$, we define
\[
\theta(\lambda) \;=\; \lambda\,\theta_{\text{base}} \;+\; (1-\lambda)\,\theta_{\text{ckpt}},
\]
so that $\lambda{=}1$ recovers the base model and $\lambda{=}0$ recovers the trained checkpoint.
At each $\lambda$ on a fixed grid, we evaluate AIME24 (pass@1, 64-run average). A smooth/high
trajectory indicates a flatter, more robust region with fewer barriers; a sharp/erratic trajectory
suggests a bumpier landscape and stronger interference among supervision signals.

\textbf{Findings.}
On both \textsc{BOBA-200} and \textsc{S1K-200} with the 8B student, MoT yields a \textbf{much smoother}
and more stable performance curve than MTD as $\lambda$ varies: performance rises steadily toward the
checkpoint and decays gradually away from it. This behavior is consistent with MoT training in a
\textbf{flatter} region (greater robustness to weight perturbations) and \textbf{better reconciliation of
cross-teacher supervision conflicts}. In contrast, MTD exhibits steeper drops and local irregularities,
implying residual inter-teacher interference.

\begin{figure}[H]
\centering
\begin{minipage}[t]{0.485\linewidth}
\centering
\includegraphics[width=\linewidth]{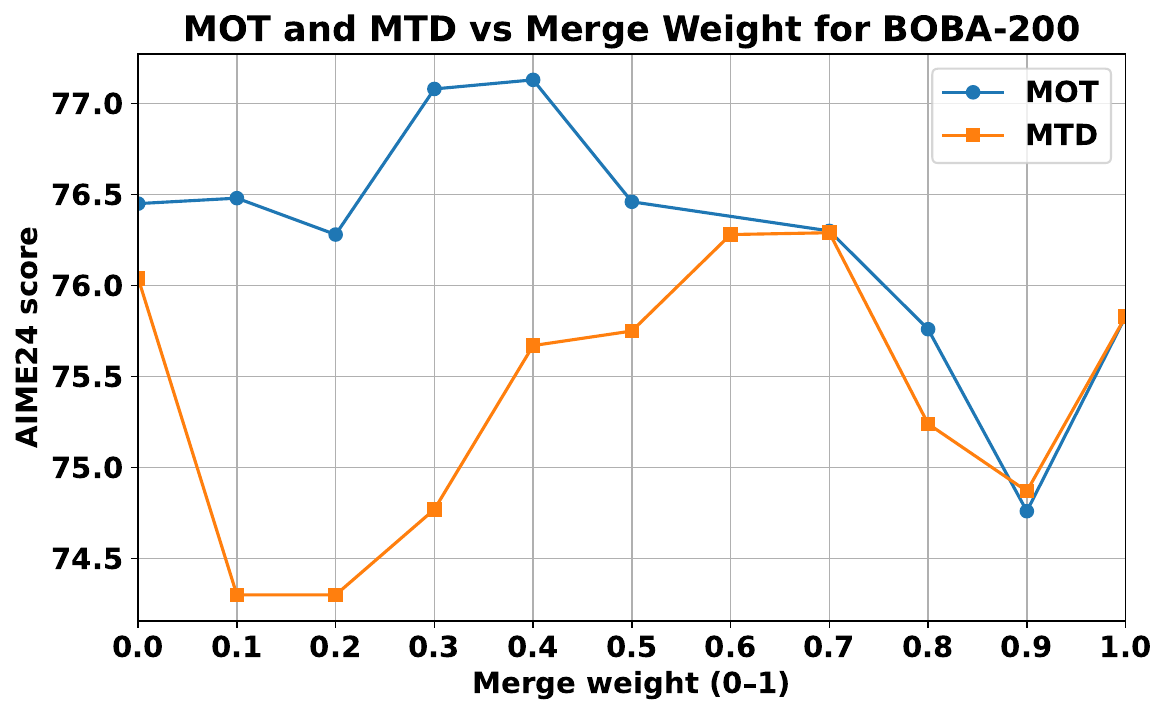}
\\[-2pt]
{\footnotesize (a) Qwen3-8B on \textsc{BOBA-200}.}
\end{minipage}
\hfill
\begin{minipage}[t]{0.485\linewidth}
\centering
\includegraphics[width=\linewidth]{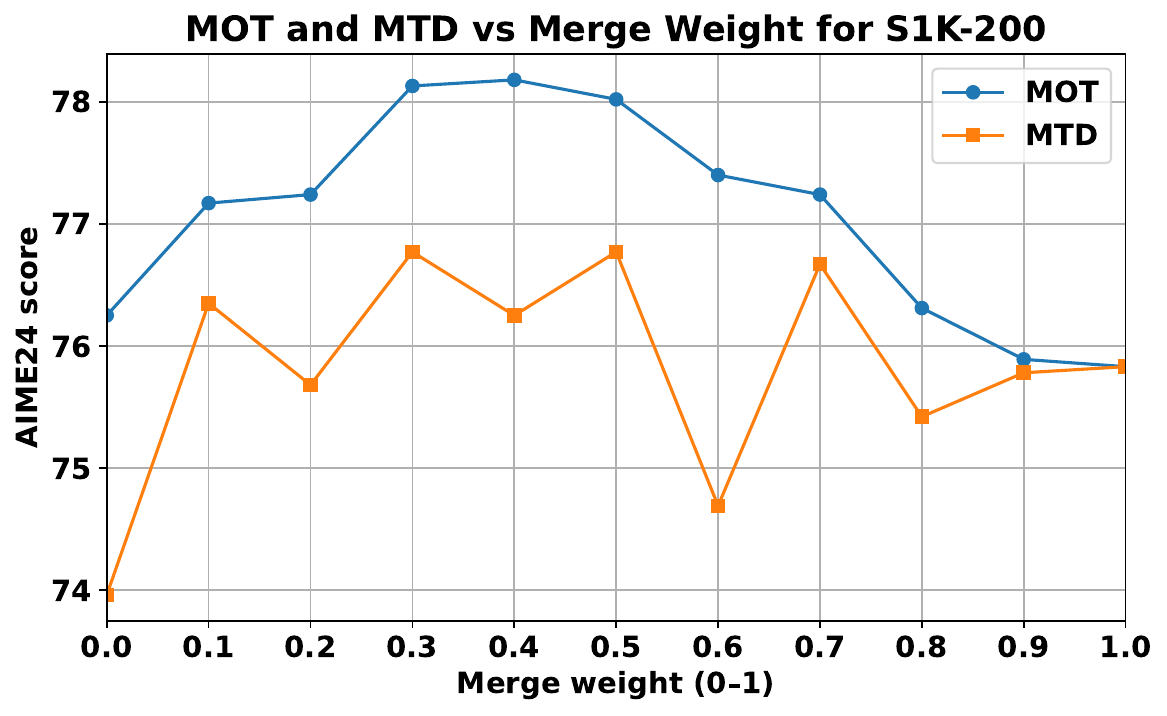}
\\[-2pt]
{\footnotesize (b) Qwen3-8B on \textsc{S1K-200}.}
\end{minipage}
\caption{Base-to-checkpoint linear interpolation (LMC). MoT shows smoother, higher trajectories than
MTD on AIME24, indicating a flatter loss region and more robust training.}
\label{fig:reverse-merge-8b}
\end{figure}

\section{Task-Type Breakdown across STD/MTD/MoT}
\label{app:Task-Type Breakdown}

\noindent\textbf{Setup and goal.}
We provide a consolidated evaluation on \textbf{BOBA-200} across \emph{all} STD/MTD settings alongside MoT, covering nine benchmarks:
catastrophic-forgetting–sensitive (CEV/SG/IFE), reasoning-knowledge (SQ/MP/MR), and pure reasoning (PB/LCB/GPQA-D).
For each setting, we report raw scores and \emph{group-wise average changes} versus the same-scale Base:
“Avg drop (cat.)” for catastrophic-forgetting–sensitive tasks (negative indicates a drop),
“Avg gain (reason.)” for reasoning-knowledge tasks, and
“Avg gain (pure)” for pure reasoning tasks. We observe a trade-off among STD choices (stronger reasoning vs.\ better forgetting mitigation), while \textbf{MoT} simultaneously yields strong math/general reasoning gains and \emph{significantly} mitigates catastrophic forgetting.

\noindent\textbf{Summary.}
Results are shown in Table~\ref{tab:app-boba-cat}, Table~\ref{tab:app-boba-reason} and Table~\ref{tab:app-boba-pure}. Single-teacher choices present a clear trade-off: some teachers maximize reasoning gains but induce larger average drops on forgetting-sensitive tasks, while others better preserve foundational abilities but yield smaller reasoning gains. \textbf{MoT} alleviates this tension: it delivers strong improvements on reasoning-knowledge and pure reasoning benchmarks, while reducing average drops on forgetting-sensitive tasks across scales.

\begin{table}[H]
\caption{Catastrophic-forgetting–sensitive tasks on BOBA-200 (CEV / SG / IFE). “Avg drop (cat.)” is the average change vs.\ the same-scale Base (negative indicates a drop). For Qwen3-30B-A3B, SG for \textit{STD(QWQ)} is unavailable (“—”); the average uses available metrics (CEV \& IFE) and compares to Base on the same subset.}
\label{tab:app-boba-cat}
\begin{center}
\scalebox{0.9}{
\begin{tabular}{llcccc}
\toprule
\textbf{Base model} & \textbf{Setting} & \textbf{CEV} & \textbf{SG} & \textbf{IFE} & \footnotesize\textbf{Avg drop (cat.)} \\
\midrule
\multirow{7}{*}{\textbf{Qwen3-8B}}
& Base                & 83.58 & 10.51 & 83.60 & -- \\
& STD (Qwen3-32B)     & 81.35 & 10.38 & 81.34 & \footnotesize{\textcolor{purple}{$\downarrow$1.54}} \\
& STD (Qwen3-235B)    & 83.28 &  9.57 & 81.18 & \footnotesize{\textcolor{purple}{$\downarrow$1.22}} \\
& STD (QWQ)           & 83.43 &  9.97 & 81.62 & \footnotesize{\textcolor{purple}{$\downarrow$0.89}} \\
& STD (Deepseek-R1)   & 83.06 &  9.70 & 81.79 & \footnotesize{\textcolor{purple}{$\downarrow$1.05}} \\
& MTD (All Teachers)  & 83.14 & 10.04 & 82.07 & \footnotesize{\textcolor{purple}{$\downarrow$0.81}} \\
& MoT (ours)          & 83.73 & 10.15 & 82.04 & \footnotesize{\textcolor{purple}{$\downarrow$0.59}} \\
\midrule
\multirow{7}{*}{\textbf{Qwen3-14B}}
& Base                & 86.78 & 10.76 & 84.69 & -- \\
& STD (Qwen3-32B)     & 84.55 & 10.07 & 82.76 & \footnotesize{\textcolor{purple}{$\downarrow$1.62}} \\
& STD (Qwen3-235B)    & 83.73 & 10.26 & 82.56 & \footnotesize{\textcolor{purple}{$\downarrow$1.89}} \\
& STD (QWQ)           & 83.73 & 10.22 & 82.36 & \footnotesize{\textcolor{purple}{$\downarrow$1.97}} \\
& STD (Deepseek-R1)   & 84.32 &  9.92 & 82.91 & \footnotesize{\textcolor{purple}{$\downarrow$1.69}} \\
& MTD (All Teachers)  & 85.14 &  9.88 & 82.22 & \footnotesize{\textcolor{purple}{$\downarrow$1.66}} \\
& MoT (ours)          & 86.70 & 10.38 & 83.51 & \footnotesize{\textcolor{purple}{$\downarrow$0.55}} \\
\midrule
\multirow{7}{*}{\textbf{Qwen3-30B-A3B}}
& Base                & 85.88 & 10.66 & 83.76 & -- \\
& STD (Qwen3-32B)     & 85.74 &  9.93 & 82.32 & \footnotesize{\textcolor{purple}{$\downarrow$0.77}} \\
& STD (Qwen3-235B)    & 84.18 & 10.02 & 80.44 & \footnotesize{\textcolor{purple}{$\downarrow$1.89}} \\
& STD (QWQ)           & 83.80 &  9.65 & 80.03 & \footnotesize{\textcolor{purple}{$\downarrow$2.27}} \\
& STD (Deepseek-R1)   & 83.36 &  9.31 & 80.61 & \footnotesize{\textcolor{purple}{$\downarrow$2.34}} \\
& MTD (All Teachers)  & 84.55 & 10.12 & 79.77 & \footnotesize{\textcolor{purple}{$\downarrow$1.95}} \\
& MoT (ours)          & 86.55 & 10.52 & 83.54 & \footnotesize{\textcolor{teal}{$\uparrow$0.10}} \\
\bottomrule
\end{tabular}
}
\end{center}
\end{table}

\begin{table}[H]
\caption{Reasoning-related tasks on BOBA-200 (SQ / MP / MR). “Avg gain (reason.)” is the average change vs.\ the same-scale Base (positive indicates an increase).}
\label{tab:app-boba-reason}
\begin{center}
\scalebox{0.9}{
\begin{tabular}{llcccc}
\toprule
\textbf{Base model} & \textbf{Setting} & \textbf{SQ} & \textbf{MP} & \textbf{MR} & \footnotesize\textbf{Avg gain (reason.)} \\
\midrule
\multirow{7}{*}{\textbf{Qwen3-8B}}
& Base                & 32.31 & 71.42 & 83.21 & -- \\
& STD (Qwen3-32B)     & 34.37 & 73.05 & 84.82 & \footnotesize{\textcolor{teal}{$\uparrow$1.77}} \\
& STD (Qwen3-235B)    & 32.63 & 72.83 & 84.84 & \footnotesize{\textcolor{teal}{$\uparrow$1.12}} \\
& STD (QWQ)           & 33.88 & 72.00 & 84.42 & \footnotesize{\textcolor{teal}{$\uparrow$1.12}} \\
& STD (Deepseek-R1)   & 33.88 & 70.92 & 84.21 & \footnotesize{\textcolor{teal}{$\uparrow$0.69}} \\
& MTD (All Teachers)  & 33.60 & 72.34 & 84.65 & \footnotesize{\textcolor{teal}{$\uparrow$1.22}} \\
& MoT (ours)          & 34.44 & 73.30 & 84.42 & \footnotesize{\textcolor{teal}{$\uparrow$1.74}} \\
\midrule
\multirow{7}{*}{\textbf{Qwen3-14B}}
& Base                & 32.61 & 75.26 & 85.74 & -- \\
& STD (Qwen3-32B)     & 32.31 & 75.36 & 85.93 & \footnotesize{\textcolor{purple}{$\downarrow$0.00}} \\
& STD (Qwen3-235B)    & 32.17 & 74.71 & 86.37 & \footnotesize{\textcolor{purple}{$\downarrow$0.12}} \\
& STD (QWQ)           & 32.42 & 74.76 & 85.19 & \footnotesize{\textcolor{purple}{$\downarrow$0.41}} \\
& STD (Deepseek-R1)   & 32.63 & 74.04 & 86.04 & \footnotesize{\textcolor{purple}{$\downarrow$0.30}} \\
& MTD (All Teachers)  & 32.77 & 74.97 & 85.82 & \footnotesize{\textcolor{purple}{$\downarrow$0.02}} \\
& MoT (ours)          & 32.65 & 75.59 & 86.53 & \footnotesize{\textcolor{teal}{$\uparrow$0.39}} \\
\midrule
\multirow{7}{*}{\textbf{Qwen3-30B-A3B}}
& Base                & 31.68 & 75.26 & 85.81 & -- \\
& STD (Qwen3-32B)     & 32.26 & 76.12 & 86.46 & \footnotesize{\textcolor{teal}{$\uparrow$0.70}} \\
& STD (Qwen3-235B)    & 31.52 & 75.96 & 86.04 & \footnotesize{\textcolor{teal}{$\uparrow$0.26}} \\
& STD (QWQ)           & 32.24 & 75.28 & 84.86 & \footnotesize{\textcolor{purple}{$\downarrow$0.12}} \\
& STD (Deepseek-R1)   & 33.00 & 72.55 & 84.16 & \footnotesize{\textcolor{purple}{$\downarrow$1.01}} \\
& MTD (All Teachers)  & 32.31 & 74.75 & 86.67 & \footnotesize{\textcolor{teal}{$\uparrow$0.33}} \\
& MoT (ours)          & 32.26 & 76.21 & 86.74 & \footnotesize{\textcolor{teal}{$\uparrow$0.82}} \\
\bottomrule
\end{tabular}
}
\end{center}
\end{table}

\begin{table}[H]
\centering
\caption{Pure reasoning tasks on BOBA-200 (PB / LCB / GPQA-D). “Avg gain (pure)” is the average change vs.\ the same-scale Base (positive indicates an increase).}
\label{tab:app-boba-pure}
\begin{center}
\scalebox{0.9}{
\begin{tabular}{llcccc}
\toprule
\textbf{Base model} & \textbf{Setting} & \textbf{PB} & \textbf{LCB} & \textbf{GPQA-D} & \footnotesize\textbf{Avg gain (pure)} \\
\midrule
\multirow{7}{*}{\textbf{Qwen3-8B}}
& Base                & 20.47 & 55.76 & 57.77 & -- \\
& STD (Qwen3-32B)     & 23.19 & 59.06 & 57.42 & \footnotesize{\textcolor{teal}{$\uparrow$1.89}} \\
& STD (Qwen3-235B)    & 23.17 & 57.90 & 58.11 & \footnotesize{\textcolor{teal}{$\uparrow$1.73}} \\
& STD (QWQ)           & 22.85 & 59.88 & 59.85 & \footnotesize{\textcolor{teal}{$\uparrow$2.86}} \\
& STD (Deepseek-R1)   & 21.90 & 56.78 & 56.50 & \footnotesize{\textcolor{teal}{$\uparrow$0.39}} \\
& MTD (All Teachers)  & 22.47 & 54.79 & 60.32 & \footnotesize{\textcolor{teal}{$\uparrow$1.19}} \\
& MoT (ours)          & 24.07 & 58.79 & 60.54 & \footnotesize{\textcolor{teal}{$\uparrow$3.13}} \\
\midrule
\multirow{7}{*}{\textbf{Qwen3-14B}}
& Base                & 28.53 & 61.41 & 60.83 & -- \\
& STD (Qwen3-32B)     & 30.72 & 62.84 & 61.52 & \footnotesize{\textcolor{teal}{$\uparrow$1.44}} \\
& STD (Qwen3-235B)    & 30.61 & 63.21 & 63.79 & \footnotesize{\textcolor{teal}{$\uparrow$2.28}} \\
& STD (QWQ)           & 28.36 & 62.80 & 63.44 & \footnotesize{\textcolor{teal}{$\uparrow$1.28}} \\
& STD (Deepseek-R1)   & 27.29 & 61.15 & 62.91 & \footnotesize{\textcolor{teal}{$\uparrow$0.19}} \\
& MTD (All Teachers)  & 29.51 & 58.50 & 63.19 & \footnotesize{\textcolor{teal}{$\uparrow$0.14}} \\
& MoT (ours)          & 30.77 & 63.59 & 64.26 & \footnotesize{\textcolor{teal}{$\uparrow$2.62}} \\
\midrule
\multirow{7}{*}{\textbf{Qwen3-30B-A3B}}
& Base                & 28.57 & 61.08 & 59.76 & -- \\
& STD (Qwen3-32B)     & 33.43 & 61.79 & 60.48 & \footnotesize{\textcolor{teal}{$\uparrow$2.10}} \\
& STD (Qwen3-235B)    & 33.31 & 61.34 & 61.81 & \footnotesize{\textcolor{teal}{$\uparrow$2.35}} \\
& STD (QWQ)           & 32.44 & 60.74 & 60.32 & \footnotesize{\textcolor{teal}{$\uparrow$1.36}} \\
& STD (Deepseek-R1)   & 29.31 & 59.02 & 59.66 & \footnotesize{\textcolor{purple}{$\downarrow$0.47}} \\
& MTD (All Teachers)  & 32.50 & 56.85 & 61.33 & \footnotesize{\textcolor{teal}{$\uparrow$0.42}} \\
& MoT (ours)          & 33.46 & 62.54 & 62.34 & \footnotesize{\textcolor{teal}{$\uparrow$2.98}} \\
\bottomrule
\end{tabular}
}
\end{center}
\end{table}

\section{Dataset}

\begin{table}[H]
\caption{STD and MTD distillation datasets derived from BOBA-200 and S1K-200.}
\begin{center}
\begin{tabular}{llll}
\toprule
Source & Teacher & Distillation dataset name & Size \\
\midrule
\multirow{5}{*}{BOBA-200} & QWQ         & BOBA-200-QWQ      & 195 \\
                          & Qwen3-32B   & BOBA-200-32B      & 191 \\
                          & Qwen3-235B  & BOBA-200-235B     & 197 \\
                          & Deepseek-R1 & BOBA-200-R1       & 198 \\
                          & ALL TEACHERS & BOBA-200-MTD & 781 \\
\midrule
\multirow{5}{*}{S1K-200}  & QWQ         & S1K-200-QWQ       & 161 \\
                          & Qwen3-32B   & S1K-200-32B       & 164 \\
                          & Qwen3-235B  & S1K-200-235B      & 169 \\
                          & Deepseek-R1 & S1K-200-R1        & 168 \\
                          & ALL TEACHERS & S1K-200-MTD  & 662 \\
\bottomrule
\end{tabular}
\label{tab:distill-all}
\end{center}
\end{table}

\section{Additional training details and full ablations}
\label{app:ablation1}

\subsection{Training hyperparameters}
\label{hyperparameters}
Unless otherwise noted, all experiments follow a shared set of training choices designed for long chain-of-thought (CoT) sequences and stable optimization:
\begin{itemize}
\item Model/input formatting: We use the Qwen3 instruction template to format prompts and responses consistently across datasets.
\item Context length: The maximum sequence length is 25k tokens to accommodate long CoT traces with minimal truncation.
\item Precision and kernels: Training uses bfloat16 with FlashAttention-2 to improve memory efficiency and throughput for long contexts.
\item Optimizer and schedule: AdamW with betas (0.9, 0.95), weight decay 0.1, cosine learning-rate schedule with a base learning rate of 1e-5 and 1\% warmup. Gradients are clipped at a norm of 1.0 for stability.
\item Batch and accumulation: We train on 8$\times$ H800 GPUs with a per-device batch size of 1 and gradient accumulation of 8, resulting in an effective batch size of 64 sequences per optimization step.
\item Logging and checkpointing: We log every step and save a checkpoint every 50 steps; up to 10 most recent checkpoints are kept, and only model weights are saved to reduce I/O overhead.
\end{itemize}

Protocol-specific details:
\begin{itemize}
\item MoT: One ``round'' consists of 50 optimization steps on a given teacher corpus before merging; we run five rounds and evaluate after each merge.
\item STD/MTD: We train for 250 steps and save/evaluate checkpoints every 50 steps; the best checkpoint is reported in the main text.
\end{itemize}

\subsection{STD/MTD and MoT per-checkpoint results}
\label{app:MoT}
For STD and MTD, we train for 250 steps and save a checkpoint every 50 steps; we evaluate each checkpoint and report the best in the main text. 

For MoT, we alternate the base model across the four STD corpora (QWQ, Qwen3-32B, Qwen3-235B, Deepseek-R1), training 50 steps on each corpus and then performing a merge; this constitutes one merge round. We run five rounds in total and evaluate after every round. The complete per-round results for all base models and both sources (BOBA-200 and S1K-200) are reported in Table~\ref{tab:appendix-ablations}.

Key observations from the ablations:

1. MoT consistently yields the strongest distillation gains in almost all settings.

2. For 8B/14B bases, MTD typically surpasses the best single-teacher STD, indicating beneficial complementarity across teachers.

3. For 30B-A3B, MTD brings little to no gain. We hypothesize that QWQ, Qwen3-32B, and Deepseek-R1 are not clearly stronger than the 30B base, so the union is dominated by Qwen3-235B; in contrast, MoT can glean useful signals from the other teachers while mitigating noise, yielding the best results.

\begin{table*}[t]
\setlength{\tabcolsep}{3.2pt}
\caption{Complete ablations on AIME 2024 (A24) and AIME 2025 (A25). Each entry is a 16-run average. We report per-checkpoint results for STD/MTD (every 50 steps, up to 250), and per-round results for MoT (Rounds 1–5).}
\label{tab:appendix-ablations}
\begin{center}
\resizebox{\textwidth}{!}{
\begin{tabular}{llcccccccccccc}
\toprule
\multirow{3}{*}{Method} & \multirow{3}{*}{Config} & \multicolumn{6}{c}{BOBA-200} & \multicolumn{6}{c}{S1K-200} \\
\cmidrule(lr){3-8}\cmidrule(lr){9-14}
 &  & \multicolumn{2}{c}{Qwen3-8B} & \multicolumn{2}{c}{Qwen3-14B} & \multicolumn{2}{c}{Qwen3-30B-A3B} & \multicolumn{2}{c}{Qwen3-8B} & \multicolumn{2}{c}{Qwen3-14B} & \multicolumn{2}{c}{Qwen3-30B-A3B} \\
\cmidrule(lr){3-4}\cmidrule(lr){5-6}\cmidrule(lr){7-8}\cmidrule(lr){9-10}\cmidrule(lr){11-12}\cmidrule(lr){13-14}
 &  & A24 & A25 & A24 & A25 & A24 & A25 & A24 & A25 & A24 & A25 & A24 & A25 \\
\midrule
Base model & (40k) & 75.83 & 67.08 & 79.17 & 70.00 & 81.67 & 72.50 & 75.83 & 67.08 & 79.17 & 70.00 & 81.67 & 72.50 \\
\midrule
\multirow{5}{*}{STD (Qwen3-32B)} 
& STEP 50  & 75.42 & 67.71 & 77.71 & 71.25 & 81.04 & 76.04 & 77.50 & 66.67 & 79.79 & 72.50 & 79.58 & 73.13 \\
& STEP 100 & 74.17 & 65.83 & 77.71 & 68.13 & 80.83 & 72.50 & 74.58 & 68.96 & 77.71 & 70.21 & 79.58 & 70.63 \\
& STEP 150 & 75.41 & 63.96 & 78.13 & 66.04 & 81.88 & 72.92 & 73.75 & 67.71 & 79.58 & 72.08 & 80.63 & 70.42 \\
& STEP 200 & 74.58 & 63.75 & 76.67 & 66.88 & 80.63 & 75.63 & 75.21 & 66.67 & 79.79 & 69.58 & 79.58 & 70.83 \\
& STEP 250 & 73.96 & 62.92 & 77.50 & 70.21 & 79.38 & 69.79 & 76.04 & 66.04 & 77.29 & 70.63 & 79.17 & 70.00 \\
\midrule
\multirow{5}{*}{STD (Qwen3-235B)} 
& STEP 50  & 74.58 & 67.92 & 78.13 & 74.79 & 80.00 & 78.13 & 74.38 & 68.54 & 77.92 & 72.71 & 77.92 & 75.63 \\
& STEP 100 & 73.13 & 68.33 & 79.17 & 74.79 & 81.88 & 75.42 & 72.50 & 65.83 & 77.08 & 75.41 & 77.08 & 76.88 \\
& STEP 150 & 71.88 & 66.67 & 78.13 & 70.42 & 77.92 & 76.04 & 74.17 & 67.71 & 77.71 & 72.08 & 78.54 & 74.58 \\
& STEP 200 & 71.04 & 65.83 & 77.29 & 74.17 & 79.58 & 75.83 & 71.46 & 67.29 & 78.75 & 73.13 & 78.33 & 74.58 \\
& STEP 250 & 75.00 & 67.29 & 79.38 & 74.17 & 80.42 & 73.54 & 73.96 & 67.08 & 76.67 & 71.46 & 79.17 & 76.04 \\
\midrule
\multirow{5}{*}{STD (QWQ)} 
& STEP 50  & 72.50 & 64.38 & 76.46 & 68.54 & 79.58 & 72.50 & 73.53 & 69.17 & 79.17 & 73.54 & 80.83 & 72.08 \\
& STEP 100 & 75.00 & 67.08 & 78.33 & 73.33 & 78.54 & 76.46 & 76.04 & 68.13 & 79.58 & 71.88 & 81.46 & 72.92 \\
& STEP 150 & 75.21 & 67.29 & 79.58 & 73.54 & 79.79 & 75.63 & 75.21 & 65.42 & 79.17 & 73.33 & 80.63 & 68.96 \\
& STEP 200 & 75.83 & 65.83 & 77.29 & 71.46 & 78.54 & 73.96 & 74.58 & 65.63 & 80.21 & 72.92 & 82.08 & 70.63 \\
& STEP 250 & 76.25 & 67.50 & 78.54 & 71.67 & 78.33 & 75.83 & 74.58 & 64.17 & 77.92 & 74.79 & 81.25 & 70.83 \\
\midrule
\multirow{5}{*}{STD (Deepseek-R1)} 
& STEP 50  & 67.71 & 60.21 & 74.38 & 67.50 & 78.33 & 68.96 & 70.00 & 61.46 & 73.75 & 62.92 & 78.54 & 70.63 \\
& STEP 100 & 70.21 & 53.33 & 73.75 & 63.33 & 75.00 & 69.79 & 68.54 & 58.33 & 73.33 & 63.13 & 76.46 & 64.58 \\
& STEP 150 & 65.83 & 56.04 & 74.58 & 63.75 & 74.79 & 64.38 & 67.92 & 52.08 & 73.96 & 62.71 & 75.63 & 66.04 \\
& STEP 200 & 65.21 & 53.75 & 74.58 & 64.79 & 74.58 & 67.50 & 66.67 & 55.83 & 71.88 & 61.25 & 74.17 & 65.21 \\
& STEP 250 & 66.67 & 55.42 & 72.50 & 63.54 & 75.42 & 66.88 & 66.88 & 51.67 & 72.71 & 63.96 & 74.17 & 70.00 \\
\midrule
\multirow{5}{*}{MTD (ALL TEACHERS)} 
& STEP 50  & 68.54 & 61.04 & 74.79 & 66.88 & 79.17 & 72.92 & 70.83 & 63.54 & 75.83 & 70.83 & 76.46 & 72.08 \\
& STEP 100 & 73.75 & 66.46 & 76.88 & 72.92 & 79.17 & 73.75 & 75.63 & 70.83 & 78.75 & 73.13 & 77.29 & 75.42 \\
& STEP 150 & 71.88 & 68.64 & 76.46 & 75.42 & 77.92 & 72.92 & 73.33 & 66.88 & 79.17 & 73.34 & 78.33 & 74.58 \\
& STEP 200 & 75.00 & 66.04 & 79.58 & 72.50 & 78.75 & 73.75 & 74.17 & 69.38 & 77.08 & 73.33 & 78.33 & 74.58 \\
& STEP 250 & 76.04 & 68.96 & 77.29 & 73.54 & 79.38 & 73.96 & 73.96 & 69.17 & 79.79 & 73.13 & 79.58 & 72.71 \\
\midrule
\multirow{5}{*}{MoT (ours)} 
& Round 1 & 72.29 & 66.88 & 78.75 & 73.95 & 80.63 & 73.13 & 74.79 & 69.17 & 78.33 & 69.79 & 80.00 & 75.42 \\
& Round 2 & 75.83 & 69.79 & 79.58 & 73.54 & 79.79 & 76.04 & 77.71 & 70.63 & 80.21 & 74.38 & 82.29 & 74.58 \\
& Round 3 & 76.67 & 70.42 & 80.00 & 74.79 & 80.00 & 77.92 & 76.25 & 70.00 & 80.00 & 74.38 & 79.79 & 74.79 \\
& Round 4 & 78.33 & 70.63 & 79.38 & 76.88 & 81.25 & 75.63 & 77.50 & 71.67 & 79.38 & 75.00 & 80.83 & 77.50 \\
& Round 5 & 76.45 & 66.88 & 78.96 & 73.75 & 82.92 & 78.33 & 76.25 & 68.13 & 81.67 & 75.63 & 80.00 & 77.50 \\
\bottomrule
\end{tabular}
}
\end{center}
\end{table*}

\subsection{Detailed MoT (without R1) Results on BOBA-200}
\label{app:no_r1}

Table~\ref{tab:appendix-MoT-no-r1} reports per-round AIME scores for MoT after ablating the Deepseek-R1 teacher (all other settings identical). AVG is computed as the mean of AIME24 and AIME25.

\begin{table}[H]
\caption{MoT without Deepseek-R1 on BOBA-200: per-round AIME24/AIME25 and AVG. AVG = (AIME24 + AIME25)/2.}
\label{tab:appendix-MoT-no-r1}
\begin{center}
\begin{tabular}{llccc}
\toprule
Base model & Round & AIME24 & AIME25 & AVG \\
\midrule
\multirow{5}{*}{Qwen3-8B}
& Round 1 & 75.21 & 69.17 & 72.19 \\
& Round 2 & 75.42 & 72.29 & 73.86 \\
& Round 3 & 76.67 & 70.00 & 73.34 \\
& Round 4 & 78.13 & 69.17 & 73.65 \\
& Round 5 & 76.46 & 69.79 & 73.13 \\
\midrule
\multirow{5}{*}{Qwen3-14B}
& Round 1 & 80.63 & 72.71 & 76.67 \\
& Round 2 & 79.79 & 74.58 & 77.19 \\
& Round 3 & 80.83 & 74.58 & 77.71 \\
& Round 4 & 81.04 & 74.79 & 77.92 \\
& Round 5 & 79.58 & 74.79 & 77.19 \\
\midrule
\multirow{5}{*}{Qwen3-30B}
& Round 1 & 81.88 & 75.00 & 78.44 \\
& Round 2 & 81.88 & 77.08 & 79.48 \\
& Round 3 & 81.25 & 78.75 & 80.00 \\
& Round 4 & 81.88 & 77.71 & 79.80 \\
& Round 5 & 80.42 & 80.00 & 80.21 \\
\bottomrule
\end{tabular}
\end{center}
\end{table}

Overall, while MoT without R1 remains competitive, the best AVG per model is consistently below the corresponding full MoT results reported in the main text. This supports the claim that R1 offers complementary supervision that raises the training ceiling and improves late-stage generalization.

\subsection{Detailed MoT with peer-level teachers (QWQ + Qwen3-32B) on BOBA-200}
\label{app:all_30}

We find that teacher usefulness is not limited to strictly stronger models. Although QWQ, Qwen3-32B, and Qwen3-30B-A3B have comparable parameter scale adn reasoning performance, distilling Qwen3-30B-A3B from peer-level teachers (QWQ or Qwen3-32B) still yields gains. This might imply that what truly benefits the model is not necessarily higher-quality reasoning trajectories, and reasoning trajectories distilled from peer-level teachers can still help. In addition, combining peer-level heterogeneous trajectories with MoT further improves results, and using all teachers performs best. Table~\ref{tab:peer-teacher} reports 16-run AIME averages on BOBA-200 with Qwen3-30B-A3B as the base. Table~\ref{tab:appendix-MoT-peer-qwq-32b} reports per-round AIME scores for MoT when using only peer-level teachers (QWQ and Qwen3-32B) with Qwen3-30B as the base. AVG is computed as the mean of AIME24 and AIME25.

Overall, these findings support two key conclusions: 

(1) Reasoning trajectories distilled from peer-level teachers can still help. 

(2) MoT robustly integrates complementary and even distribution-shifted supervision, extracting useful signals while mitigating noise.

\begin{table}[H]
\caption{Peer-level teachers can still help. Results on BOBA-200 with Qwen3-30B-A3B as the base; AIME scores are 16-run averages, AVG is the mean of AIME24 and AIME25.}
\label{tab:peer-teacher}
\begin{center}
\begin{tabular}{lccc}
\toprule
Teacher setting & AIME24 & AIME25 & AVG \\
\midrule
Base                             & 80.63 & 70.00 & 75.32 \\
STD: only QWQ                       & 79.79 & 75.63 & 77.71 \\
STD: only Qwen3-32B                 & 81.04 & 76.04 & 78.54 \\
MoT: QWQ + Qwen3-32B                & 81.04 & 77.29 & 79.17 \\
MoT: ALL TEACHERS                   & 82.92 & 78.33 & 80.63 \\
\bottomrule
\end{tabular}
\end{center}
\end{table}



\begin{table}[H]
\caption{MoT with peer-level teachers (QWQ + Qwen3-32B) on BOBA-200: per-round AIME24/AIME25 and AVG for Qwen3-30B. AVG = (AIME24 + AIME25)/2.}
\label{tab:appendix-MoT-peer-qwq-32b}
\begin{center}
\begin{tabular}{lccc}
\toprule
Round & AIME24 & AIME25 & AVG \\
\midrule
Round 1 & 82.70 & 73.95 & 78.33 \\
Round 2 & 80.83 & 74.58 & 77.71 \\
Round 3 & 82.08 & 75.83 & 78.96 \\
Round 4 & 80.83 & 75.00 & 77.92 \\
Round 5 & 81.04 & 77.29 & 79.17 \\
\bottomrule
\end{tabular}
\end{center}
\end{table}

\section{Benchmark Categories and Details}
\label{app:benchmarks}

We evaluate nine benchmarks under three categories—\textbf{catastrophic-forgetting–sensitive}, \textbf{reasoning–knowledge}, and \textbf{pure reasoning}—to assess whether CoT-style training with MoT preserves basic capabilities while strengthening reasoning. Here we have provided detailed content and descriptions of these tasks, and given the MoTivations for using them for evaluation and classifying them into the corresponding task categories.

\subsection{Catastrophic-Forgetting–Sensitive Tasks}

\noindent\textbf{CEVAL (CEV).}

\emph{Description:} CEVAL is a Chinese multi-discipline multiple-choice exam suite with approximately 14{,}000 items spanning 52 subjects at varying difficulty levels. 

\emph{Task:} It evaluates factual and domain knowledge recall across humanities, sciences, and professional tracks.

\emph{MoTivation:} It probes retention of broad multilingual knowledge that can degrade after CoT-style training.

\noindent\textbf{SUPER\_GPQA (SG).}

\emph{Description:} SUPER\_GPQA is a graduate-level, multi-domain multiple-choice benchmark covering a wide range of academic disciplines.

\emph{Task:} It measures advanced factual knowledge with light multi-step reasoning.

\emph{MoTivation:} It tests whether extensive pretraining knowledge is preserved following CoT fine-tuning.

\noindent\textbf{IFEVAL (IFE).}

\emph{Description:} IFEVAL is an instruction-following suite with automatically verifiable constraints such as length, formatting, and keyword usage.

\emph{Task:} It evaluates instruction compliance and adherence to explicit constraints.

\emph{MoTivation:} It checks for forgetting of fundamental alignment and compliance behaviors after CoT training.

\subsection{Reasoning–Knowledge Tasks}

\noindent\textbf{SIMPLE\_QA (SQ).}

\emph{Description:} SIMPLE\_QA is a collection of short, unambiguous fact-seeking questions with a single correct answer.
\emph{Task:} It evaluates factual accuracy and calibrated answering by discouraging uninformed guessing.

\emph{MoTivation:} It tests whether CoT improves precision while avoiding hallucinations or overconfident errors.

\noindent\textbf{MMLU\_PRO (MP).}

\emph{Description:} MMLU\_PRO is a harder variant of MMLU that increases item difficulty and option counts to emphasize reasoning.

\emph{Task:} It measures multi-step reasoning grounded in broad subject knowledge across many domains.

\emph{MoTivation:} It assesses whether CoT enhances reasoning while maintaining robust domain knowledge.

\noindent\textbf{MMLU\_REDUX (MR).}

\emph{Description:} MMLU\_REDUX is a curated and corrected subset of MMLU designed to reduce labeling noise.

\emph{Task:} It evaluates multi-subject knowledge with some analytical reasoning under cleaner annotations.

\emph{MoTivation:} It isolates capability changes from dataset artifacts and checks knowledge retention under CoT.

\subsection{Pure Reasoning Tasks}

\noindent\textbf{PhyBench (PB).}

\emph{Description:} PhyBench is a set of physics problems ranging from high-school to Olympiad level that require careful quantitative reasoning.

\emph{Task:} It measures multi-step physics reasoning including derivations and the coordination of multiple principles.

\emph{MoTivation:} It emphasizes chain-of-thought style reasoning rather than rote memorization of facts.

\noindent\textbf{LiveCodeBench (LCB).}

\emph{Description:} LiveCodeBench is a contamination-controlled suite of recent competitive programming problems drawn from diverse sources.

\emph{Task:} It evaluates algorithmic reasoning, program synthesis, and debugging under executable tests.

\emph{MoTivation:} It probes problem decomposition and step-by-step solution planning independent of encyclopedic knowledge.

\noindent\textbf{GPQA-Diamond (GPQA-D).}

\emph{Description:} GPQA-Diamond is the hardest expert-vetted subset of GPQA spanning biology, physics, and chemistry.

\emph{Task:} It measures deep scientific reasoning on challenging multiple-choice questions that resist superficial lookup.

\emph{MoTivation:} It stresses genuine multi-step reasoning and scientific insight rather than retrieval of surface facts.


\includepdf[
  pages={1-4},
  scale=0.65,
  pagecommand={
    \thispagestyle{plain}%
    \begin{center}
      \bfseries \figurename~\ref{fig:tokenconf-r1-three}. 
      Tokens marked with confidence drops in \textbf{MoT} under \textbf{R1}-distilled supervision. Marked tokens cluster on teacher-specific stylistic tokens while derivational tokens stay high. This indicates that MoT attenuates inter-teacher inductive biases while preserving consensus reasoning steps. (Note: only tokens after \textless\textbar im\_start\textbar\textgreater are included in the loss.)
    \end{center}%
  },
  pagecommand*={
    \refstepcounter{figure}\label{fig:tokenconf-r1-three}%
  }
]{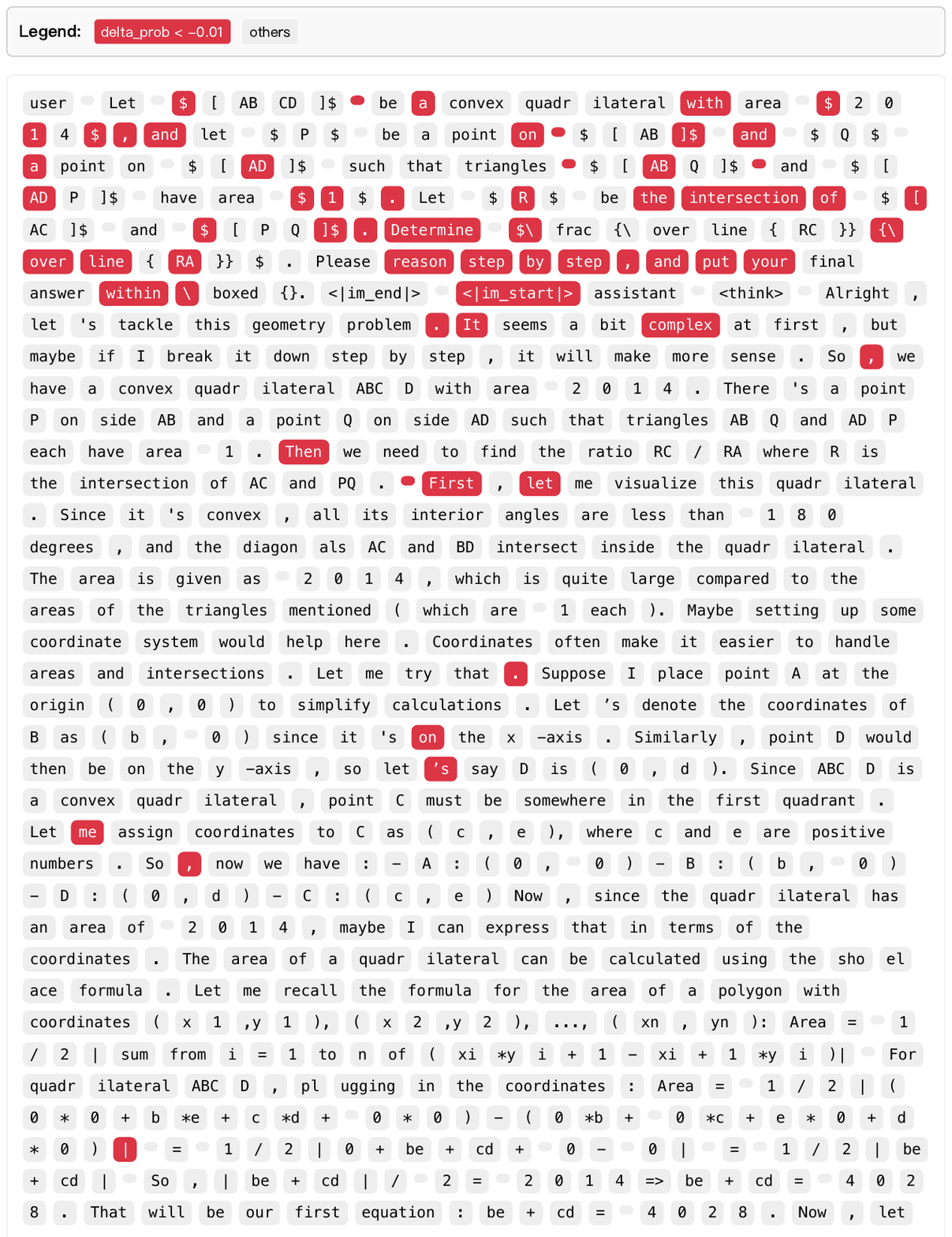}

\includepdf[
  pages={1-4},
  scale=0.65,
  pagecommand={
    \thispagestyle{plain}%
    \begin{center}
      \bfseries \figurename~\ref{fig:tokenconf-r1-std-three}. 
      Tokens marked with confidence drops in \textbf{STD(R1)} under \textbf{R1}-distilled supervision. Marked tokens are rare to nearly absent, indicating that direct distillation fully adopts the teacher’s trajectory, including stylistic tokens and latent inductive biases, rather than filtering them. (Note: only tokens after \textless\textbar im\_start\textbar\textgreater are included in the loss.)
    \end{center}%
  },
  pagecommand*={
    \refstepcounter{figure}\label{fig:tokenconf-r1-std-three}%
  }
]{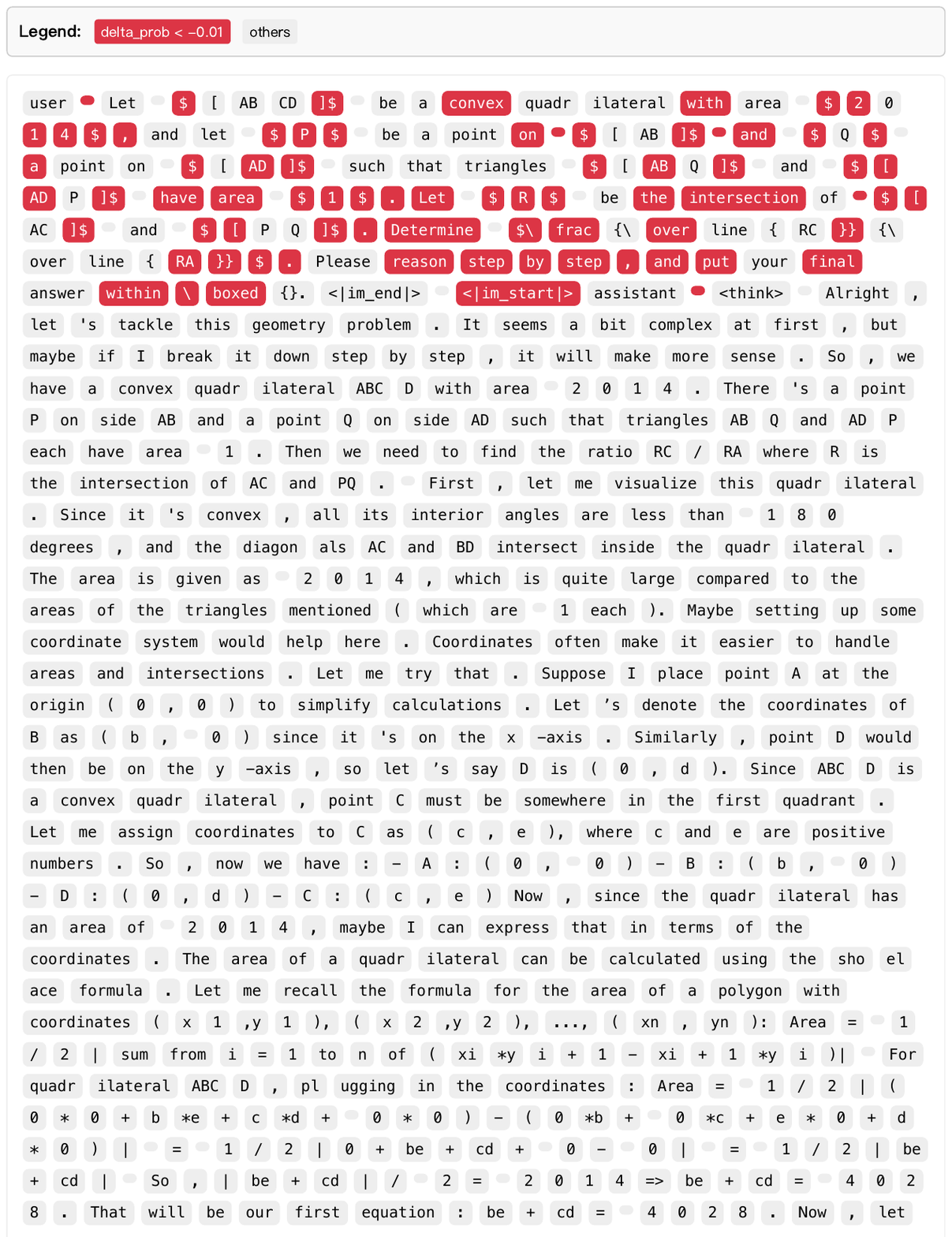}

\end{document}